\documentclass{article} % For LaTeX2e
\usepackage{iclr2026_conference,times}
\usepackage{amsmath,amsfonts,bm}

\def\eqref#1{equation~\ref{#1}}
\def\1{\bm{1}}

\DeclareMathAlphabet{\mathsfit}{\encodingdefault}{\sfdefault}{m}{sl}
\SetMathAlphabet{\mathsfit}{bold}{\encodingdefault}{\sfdefault}{bx}{n}

\usepackage{pifont}
\usepackage{hyperref}
\usepackage{graphicx}
\usepackage{url}
\usepackage{booktabs}           % \toprule \midrule \bottomrule
\usepackage[table]{xcolor}      % \rowcolor
\usepackage{makecell} 
\usepackage{amsmath} 
\usepackage{amssymb}
\usepackage{booktabs}
\usepackage{diagbox}
\usepackage[ruled,vlined,linesnumbered]{algorithm2e}

\usepackage{wrapfig}

\title{VMDiff: Visual Mixing Diffusion for Limitless Cross-Object Synthesis}

\iclrfinalcopy

\author{Zeren Xiong\textsuperscript{1} \quad Yue yu\textsuperscript{1}\quad Zedong Zhang\textsuperscript{1}\quad Shuo Chen\textsuperscript{2}\quad Jian Yang\textsuperscript{1} \quad Jun Li\textsuperscript{1}\thanks{Corresponding author: junli@njust.edu.cn} \\ \textsuperscript{1}Nanjing University of Science and Technology\quad \textsuperscript{2}Nanjing University \\ \texttt{xzr3312@gmail.com, shuo.chen@nju.edu.cn,}\\
\texttt{\{zandyz,  yuue, csjyang, junli\}@njust.edu.cn}
}

\begin{document}

\maketitle

%%%%%%%%%%%%%%%%%%%%%%%%%%%%%%%%%%%%%%%%%%%%%%%%%%%%%%%%%%%%%%%%%%%%%%%%%
\begin{figure}[h]
\begin{center}
%\framebox[4.0in]{$\;$}
\vskip -0.38in
\includegraphics[width=0.97\textwidth]{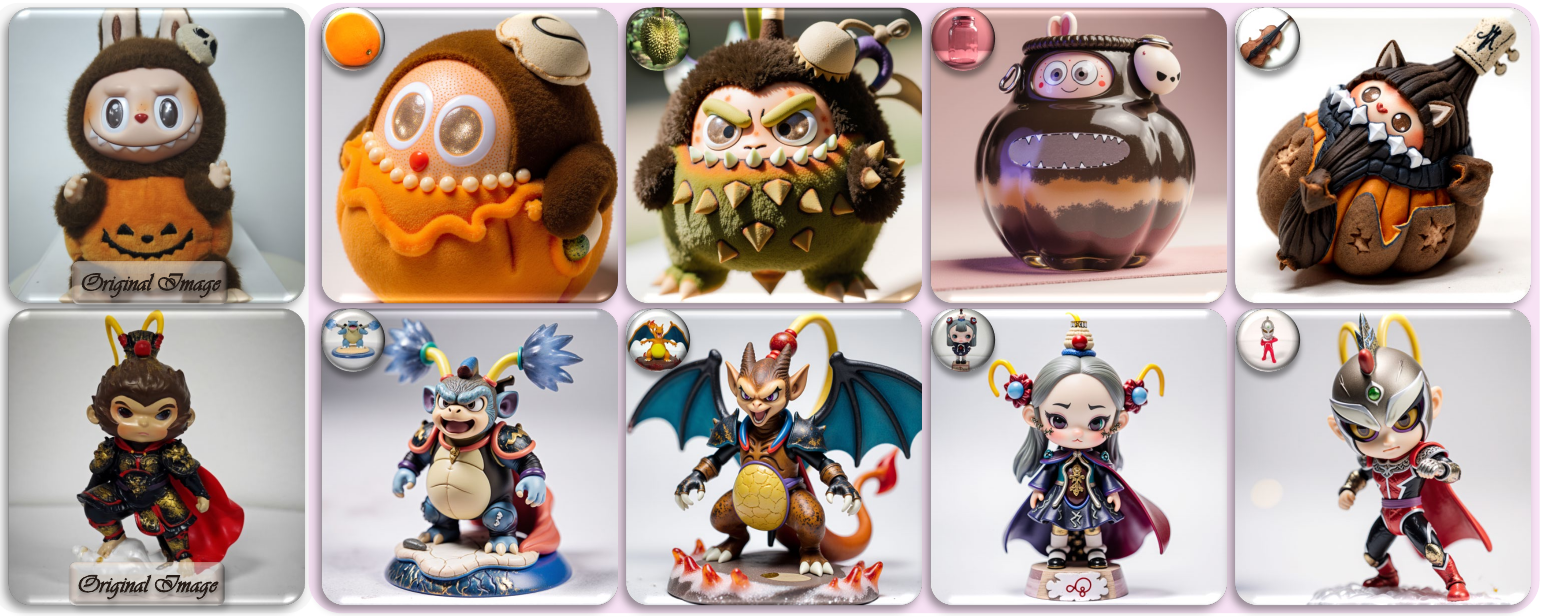}
\vspace{-0.2cm}
\end{center}
\vskip -0.1in
\caption{Two groups (rows) illustrating our VMDiff’s capability to generate coherent hybrid objects. For each group, images from the \textit{$2$nd} to the \textit{$5$th} column are the product of fusing the source image in the \textit{$1$st} column with the corresponding image in the top left\protect\footnotemark .} 
\label{fig:First_image}

\end{figure}
\footnotetext{Our method’s results, presented under the title \textit{Creative Toys Series}, were awarded the Silver Award at the NY Digital Awards 2025 (\href{https://nydigitalawards.com/winner-info.php?id=675}{\textcolor{magenta}{Award Page}}, \href{https://youtu.be/rAEvsM9uWwA}{\textcolor{magenta}{Video}}).}

%%%%%%%%%%%%%%%%%%%%%%%%%%%%%%%%%%%%%%%%%%%%%%%%%%%%%%%%%%%%%%%%%%%%%%%%%

\begin{abstract}
 Creating novel images by fusing visual cues from multiple sources is a fundamental yet underexplored problem in image-to-image generation, with broad applications in artistic creation, virtual reality and visual media. Existing methods often face two key challenges: \textit{coexistent generation}, where multiple objects are simply juxtaposed without true integration, and \textit{bias generation}, where one object dominates the output due to semantic imbalance. To address these issues, we propose \textbf{Visual Mixing Diffusion (VMDiff)}, a simple yet effective diffusion-based framework that synthesizes a single, coherent object by integrating two input images at both noise and latent levels. Our approach comprises: (1) a \textit{\textbf{hybrid sampling process}} that combines guided denoising, inversion, and spherical interpolation with adjustable parameters to achieve structure-aware fusion, mitigating coexistent generation; and (2) an \textit{\textbf{efficient adaptive adjustment}} module, which introduces a novel similarity-based score to automatically and adaptively search for optimal parameters, countering semantic bias. Experiments on a curated benchmark of 780 concept pairs demonstrate that our method outperforms strong baselines in visual quality, semantic consistency, and human-rated creativity. \href{https://xzr52.github.io/VMDiff_index/}{\textcolor{magenta}{Project}}.
\end{abstract}

%%%%%%%%%%%%%%%%%%%%%%%%%%%%%%%%%%%%%%%%%%%%%%%%%%%%%%%%%%%%%%%%%%%%
\begin{figure}[t]
  \centering
  \includegraphics[width=0.94\linewidth]{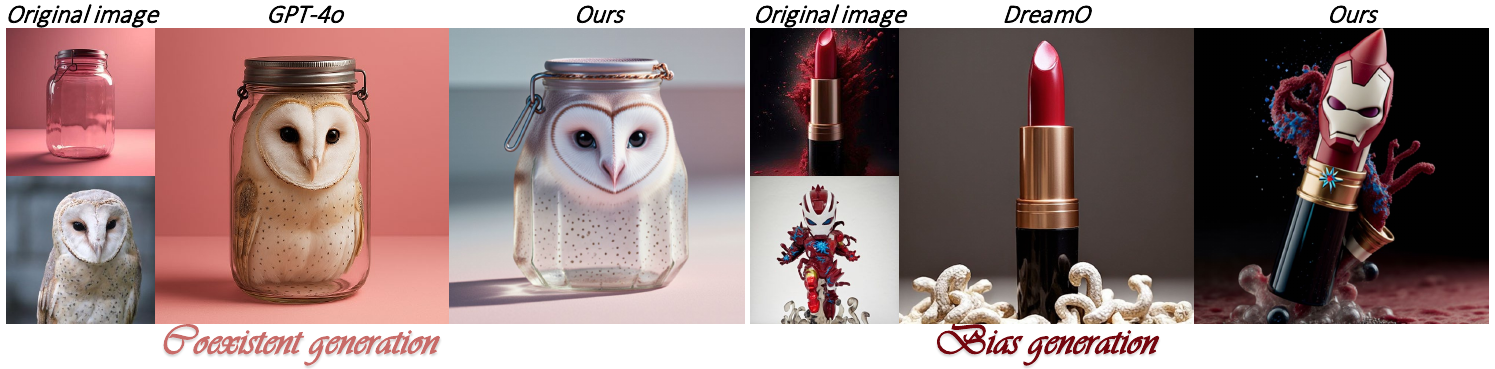}
  \vskip -0.15in
  \caption{\textbf{Failed fusions between two object images.} GPT-4o \cite{openai2023chatgpt} performs \textit{\textbf{coexistent generations}} (left), while DreamO \cite{mou2025dreamo} exhibits \textit{\textbf{bias generations}} (right). In contrast, our method achieves a seamless and harmonious fusion of the two objects.
% characteristics
  }

  \label{fig:problem}
\vskip -0.21in
\end{figure}
%%%%%%%%%%%%%%%%%%%%%%%%%%%%%%%%%%%%%%%%%%%%%%%%%%%%%%%%%%%%%%%%%%%%

\section{Introduction}
\label{sec:intro}
Synthesizing novel images by combining visual elements from multiple sources is a fundamental challenge in image-to-image generation, with wide applications in virtual reality \cite{haque2023instruct,chen2024gaussianeditor}, digital media \cite{zheng2024probing,zhao2024equivariant}, product design \cite{ju2023direct,sheynin2024emu,MoA2024} and film and game \cite{ceylan2023pix2video,liu2024video}. In particular, visual composition methods generate high-fidelity images by composing objects through various strategies, such as combining object words into complex sentences \cite{liu2022compositional}, merging multiple objects \cite{liu2021learning}, or blending scenes and styles \cite{zou2025mcig}. Although these approaches effectively position different objects or parts within an image, they often struggle to seamlessly integrate distinct elements into a single object. Recent semantic mixing \cite{li2024tp2ocreativetextpairtoobject,xiong2024novel} explores novel object synthesis by combining textual descriptions of one object with another images or text. In contrast, this work focuses on visual mixing—directly blending two object images into a single, imaginative, and visually cohesive concept.

However, when applying existing powerful methods are used to perform this visual mixing task, we identify two key limitations. First, \textbf{coexistent generation} (see Fig.~\ref{fig:problem}, left) occurs when different objects merely appear in the same scene—either side-by-side or partially overlapped—without achieving true visual and semantic integration. While the resulting compositions are spatially coherent, they remain conceptually disjoint. For example, OpenAI’s recent GPT-4o \cite{openai2023chatgpt} produces an image where the violin and pineapple overlap but fail to meaningfully fuse.
Second, \textbf{bias generation} (see Fig.~\ref{fig:problem}, right) arises when the model generates only one object while omitting the other. This asymmetry likely stems from imbalanced representations or unresolved semantic conflicts, leading to outputs that disproportionately emphasize one object. For instance, OmniGen \cite{xiao2024omnigen} generates the doll figurine while entirely neglecting the horse.

To address these limitations, we develop \textbf{Visual Mixing Diffusion (VMDiff)}, a simple yet effective framework for synthesizing novel, coherent objects that seamlessly integrate two input images. VMDiff ensures structural plausibility and semantic balance through two key components: a \textbf{\textit{Hybrid Sampling Process (HSP)}} and an \textbf{\textit{Efficient Adaptive Adjustment (EAA)}}. HSP integrates the two inputs through noise inversion and feature fusion. The inversion refines an initial noise vector conditioned on a concatenated input object embedding with two parameters and their corresponding text prompt, ensuring deep information mixing to prevent mere juxtaposition. Subsequently, feature fusion employs a curvature-respecting interpolation to blend image embeddings, with a scale factor controlling either object from dominating and thus countering bias generation. EAA automates the search for optimal parameters by proposing a novel similarity-based score that measures alignment with both visual/semantic similarity and balance between the fused object and the input object images/their category labels. By maximizing this score, the EAA dynamically adjusts the influence of each input, ensuring semantically coherent and visually faithful fusions across diverse object pairs.
%\textit{(e.g., A photo of $<T_1>$ creatively fused with $<T_2>$)}

Our contributions are summarized as follows: \textbf{(1)} We introduce a \textit{hybrid sampling process} that constructs optimized semantic noise via guided denoising and inversion, combined with a curvature-aware latent fusion strategy using spherical interpolation for smooth and tunable blending.
\textbf{(2)} We present an \textit{efficient adaptive adjustment} algorithm that adjusts fusion parameters to achieve semantic and visual balance via a lightweight score-driven search. 
\textbf{(3)} By integrating them, we propose VMDiff, a unified and controllable framework for object-level visual concept fusion. Experiments on a curated benchmark of 780 concept pairs demonstrate that our method achieves superior object synthesis, excelling in semantic consistency, visual harmony, and user-rated creativity.

\section{Related Work}

% Head 2
\textbf{Multi-Concept Generation.} 
Multi-concept generation seeks to synthesize images representing multiple user-defined concepts, typically from a few reference images per concept. Early works such as Custom Diffusion~\cite{kumari2023multi} and SVDiff~\cite{han2023svdiff} extend single-concept personalization by fine-tuning on joint data or merging customized models. Later methods~\cite{ gu2023mix, liu2023cones} enhance compositionality by merging LoRA modules or token embeddings via gradient fusion~\cite{gu2023mix} or spatial inversion~\cite{zhang2024compositional}. 
More recent approaches further improve efficiency and flexibility: FreeCustom~\cite{ding2024freecustom} employs multi-reference self-attention and weighted masks for training-free composition, while MIP-Adapter~\cite{huang2025resolving} mitigates object confusion with a weighted-merge strategy. OmniGen~\cite{xiao2024omnigen} and DreamO~\cite{mou2025dreamo} provide unified instruction-based frameworks for diverse generation tasks. 
Unlike prior methods that explicitly separate input concepts, our approach introduces a unified fusion framework that integrates two concept inputs into a novel object with coherent structure and balanced semantics.

% Head 3
\textbf{Semantic Mixing.} 
Creativity, spanning domains from scientific theories to culinary recipes, has long been a key driver of progress in artificial intelligence~\cite{boden2004creative,maher2010evaluating,wang2023creative,xiong2025category}. In this context, semantic mixing has emerged as a promising approach for generating novel objects by fusing features from multiple concepts into a single coherent representation. Unlike traditional style transfer~\cite{Zhang2023inversion,Tang2023master,ke2023neural} or image editing~\cite{avrahami2024stable,Dong2023prompt,brooks2023instructpix2pix,Gal2023personalizeT2I}—which emphasize texture transfer or localized modifications while preserving layout—semantic mixing focuses on concept-level integration within a single entity. 
Conceptlab~\cite{Richardson2024conceptlab} interpolates token embeddings to synthesize imaginative entities, while TP2O~\cite{li2024tp2ocreativetextpairtoobject} enhances controllability by aligning and blending prompt embeddings. However, both operate purely in the textual domain and lack support for real visual content. 
MagicMix~\cite{Liew2022Magicmix} fuses image latents with text prompts during denoising, preserving spatial structure, while ATIH~\cite{xiong2024novel} improves semantic alignment through more coordinated integration of visual and textual inputs. FreeBlend~\cite{zhou2025freeblend} performs staged interpolation in latent space to produce blended objects. 
In contrast, our method integrates structural and semantic cues from real image concepts, generating hybrid objects that are both visually coherent and semantically balanced.

\section{Visual Mixing Diffusion}
\begin{figure}[t]
  \centering
  \includegraphics[width=0.97\linewidth]{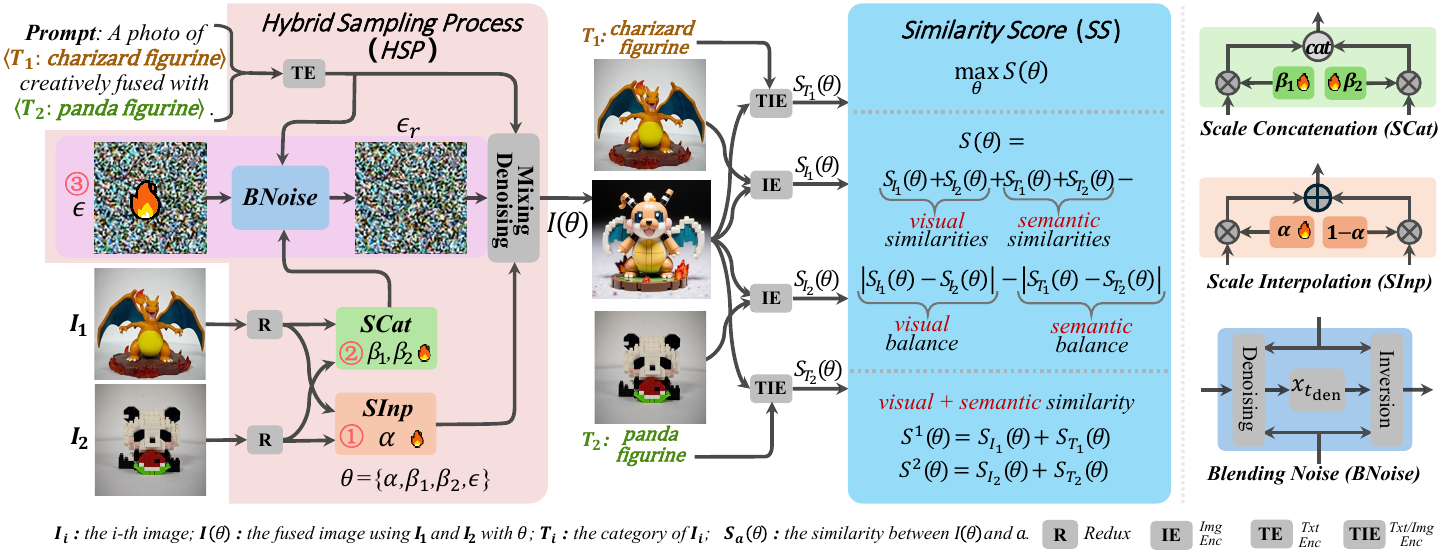}
  \vskip -0.15in
  \caption{\textbf{Overview of our VMDiff framework.} Given two input images and their categories, the Hybrid Sampling Process (HSP) fuses them using noise inversion, scale interpolation (SInp) and scale concatenation (SCat). Efficient adaptive adjustment (EAA) optimizes fusion parameters \(\theta = \{\alpha, \beta_1, \beta_2, \epsilon\}\) via a similarity score (SS) that measures visual, semantic, and balance consistency.} %The final 
  \label{fig:pipeline}
  \vskip -0.1in
\end{figure}

In this section, we present a Visual Mixing Diffusion (\textbf{VMDiff}) for synthesizing novel objects images in Fig.~\ref{fig:pipeline}. Our method consists of two key components. We introduce a Hybrid Sampling Process (\textbf{HSP, §\ref{sec:vmdiff}}) that generates a new object image by blending two distinct inputs using learned scale factors and noise. An Efficient Adaptive Adjustment (\textbf{EAA, §\ref{sec:EAA}}) dynamically adjusts the scale factors and noise based on a Similarity Score (\textbf{SS}), ensuring high-quality object synthesis. 

\subsection{Hybrid Sampling Process}
\label{sec:vmdiff}

Given two distinct images $I_1$ and $I_2$, along with their respective category labels $T_1$ and $T_2$ (e.g., \textit{Iron Man} and \textit{Duck}), we first construct a guiding prompt $P_{G}$: “\textit{A photo of $<T_1>$ creatively fused with $<T_2>$.}” and sample an initial Gaussian noise $\epsilon\sim \mathcal{N}(0,I)$. For convenience, we denote an input data $D=\{I_1,I_2,T_1,T_2,P_G\}$. We first employ pretrained image/text encoders $\mathcal{E}_I(\cdot)/ \mathcal{E}_T(\cdot)$ of FLUX-Krea~\cite{flux1kreadev2025} to project both visual and textual modalities into a unified image-language latent space. Specifically, these embeddings are extracted by
\(z_1 = \mathcal{E}_I(I_1), \ z_2 = \mathcal{E}_I(I_2), \ z_{p} = \mathcal{E}_T(P_G)\).
Using these embeddings, HSP includes \textit{blending noise} and \textit{mixing denoise}. 

\begin{wrapfigure}{r}{0.59\textwidth}
\vskip -0.15in
  \centering
  \includegraphics[width=1\linewidth]{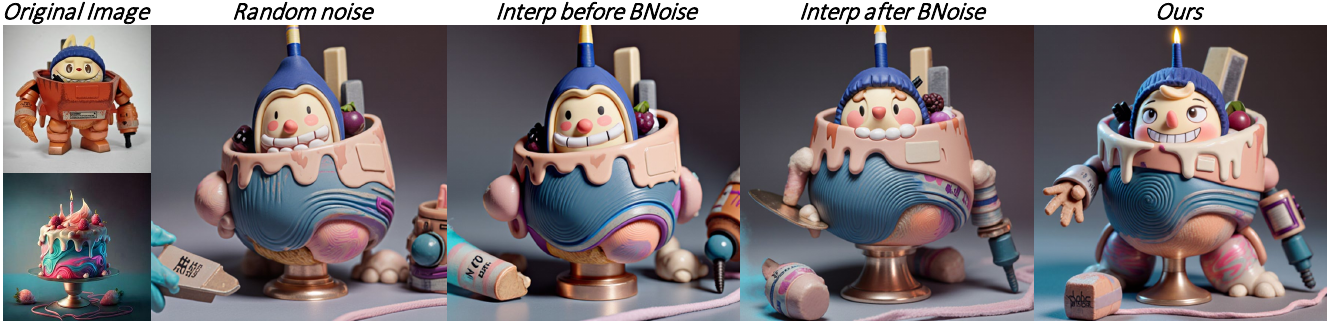}
  \vskip -0.1in
  \caption{Different BNoise strategies. } 
  \label{fig:AB_noise_refine}
   \vskip -0.15in
\end{wrapfigure}

\textbf{Blending Noise (BNoise):} Directly sampling standard Gaussian noise to generate a blend of two objects frequently produces incomplete results, with key features such as arms or legs missing (Fig. \ref{fig:AB_noise_refine}). This occurs because random noise contains no information about the input objects. Our solution is to refine an initial noise vector $\epsilon$, transforming it into a visually and semantically-informed estimate that faithfully represents the source data. Inspired by Rectified Flow~\cite{albergo2022building}, this is achieved through a guided denoising and inversion process.
Using inputs \(\epsilon,z_1,z_2,z_p\), we denoise to an intermediate timestep $t_{\text{den}}$, and invert to a refined noise \(\epsilon_b\), which is defined as:
\begin{align}
\begin{aligned}
\hat{x}_t = x_{t_{\text{den}}}&\Leftarrow \overbrace{x_{t-1} = x_t - (\sigma_t - \sigma_{t-1})v_{\phi}(x_t, t, z_{\text{SCat}}(z_1,z_2;\beta_1, \beta_2), \gamma_{\text{den}}, z_p)}^{\text{denoise:} \ t \ \text{decreases from}\ T\ \text{to}\ t_{\text{den}},\ \text{starting}\ x_T = \epsilon}, \\
\epsilon_{b}=\underbrace{\hat{x}_T}_{\text{BNoise}}&\Leftarrow \underbrace{\hat{x}_{t+1} = \hat{x}_{t} + (\sigma_{t+1} - \sigma_{t}) v_{\phi}(\hat{x}_{t}, t, z_{\text{SCat}}(z_1,z_2;\beta_1, \beta_2), \gamma_{\text{inv}}, z_p)}_{\text{inversion:} \ t \ \text{increases from}\ t_{\text{den}}\ \text{to}\ T,\ \text{starting}\ \hat{x}_t = x_{t_{\text{den}}}},
\end{aligned}
\label{eq:BNoise}
\end{align}
where \( x_{t}\) and \(\hat{x}_{t}\) are latent variables at timestep $t$, \( v_{\phi} \) denotes the noise prediction network, \(\sigma_{t}\) controls the sampler parameter. For conditioning, we adopt parameters from \cite{bai2024zigzag}: a high denoising strength \(\gamma_{\text{den}}=5\) ensures strong guidance, while an inversion strength of \(\gamma_{\text{inv}}=0\) is used to reduce distortion in the noise space. The total number of timesteps 
\(T\) is \(999\), with a predefined intermediate denoising timestep at \(t_{\text{den}}=652\). In \eqref{eq:BNoise}, \(z_p\) provides the semantic information, while \(z_{\text{SCat}}\) provides visual information. Here, we introduce two learnable factors \( \beta_1, \beta_2 \in \mathbb{R}_+ \) to create a \textbf{\textit{scale concatenation (SCat)}} of the input latents: \(z_{\text{SCat}}(z_1,z_2;\beta_1, \beta_2) = \text{concat}(\beta_1 z_1, \beta_2 z_2)\).

%%%%%%%%%%%%%%%%%%%%%%%%%%%%%%%%%%%%%%%%%%%%%%%%%%%%%%%%%%%%%%%%%%%%%%%%%
%%%%%%%%%%%%%%%%%%%%%%%%%%%%%%%%%%%%%%%%%%%%%%%%%%%%%%%%%%%%%%%%%%%%%%%%%
\textit{\textbf{Discussion on BNoise: concatenate vs. interpolate.}} We hypothesize that interpolating mismatched embeddings obscures subtle features, while concatenation preserves them, allowing the inversion process to refine noise containing the full concept. To test this, we compare \textit{Interpolate before BNoise:} Blend embeddings first, then refine the noise, and \textit{Interpolate after BNoise:} Refine noise from each embedding first, then blend the results. Fig.~\ref{fig:AB_noise_refine} shows that both interpolation methods fail to capture intricate details (e.g., legs), whereas our concatenation yields superior visual quality and faithfulness by preserving input details and ensuring a coherent denoising pathway. \textit{\textbf{Quantitative results in Appdx.~\ref{sec:Discuss}}}.

\textbf{Mixing Denoise (MDeNoise):} Using the blended noise \(\epsilon_b\), we denoise it to finally produces a cross-object fusion by mixing the inputs, \(z_1,z_2,z_p\). Specifically, we formulate this process as:
\begin{align}
I=\mathcal{D}(x_0),\ \text{where}\
x_0\Leftarrow \overbrace{x_{t-1} = x_t - (\sigma_t - \sigma_{t-1})v_{\phi}(x_t, t, z_{\text{SInp}}(z_1,z_2;\alpha), \gamma_{\text{gen}}, z_p)}^{\text{MDeNoise:} \ t \ \text{decreases from}\ T\ \text{to}\ 0,\ \text{starting}\ x_T = \epsilon_b}.
\label{eq:MDeNoise}
\end{align}
Here, \( \gamma_{\text{gen}} = 4.0 \) is a fixed guidance scale, and the decoder \(\mathcal{D}(\cdot)\) generate the final fusion image \(I\) using the FLUX-Krea decoder~\cite{flux1kreadev2025}. The \textbf{\textit{scale interpolation (SInp)}}, \(z_{\text{SInp}}(z_1,z_2;\alpha)\), mixes the two visual embeddings \( z_1 \) and \( z_2 \) into a single coherent representation, which is implemented by a spherical interpolation ~\cite{shoemake1985animating}: \(z_{\text{SInp}}(\alpha) = \tfrac{\sin(\alpha \cdot \delta)}{\sin(\delta)} z_1 + \tfrac{\sin((1 - \alpha) \cdot \delta)}{\sin(\delta)} z_2\),  
where $\delta = \cos^{-1}(z_1 \cdot z_2)$, and $0\leq\alpha\leq 1$ is a learnable factor to control the mixing ratio. This MDeNoise process in \eqref{eq:MDeNoise} outputs the final fusion image \(I\). 

%%%%%%%%%%%%%%%%%%%%%%%%%%%%%%%%%%%%%%%%%%%%%%%%%%%%%%%%%%%%%%%%%%%%%%%%%
\begin{wrapfigure}{r}{0.55\textwidth}
 \vskip -0.15in
  \centering
\includegraphics[width=0.97\linewidth]{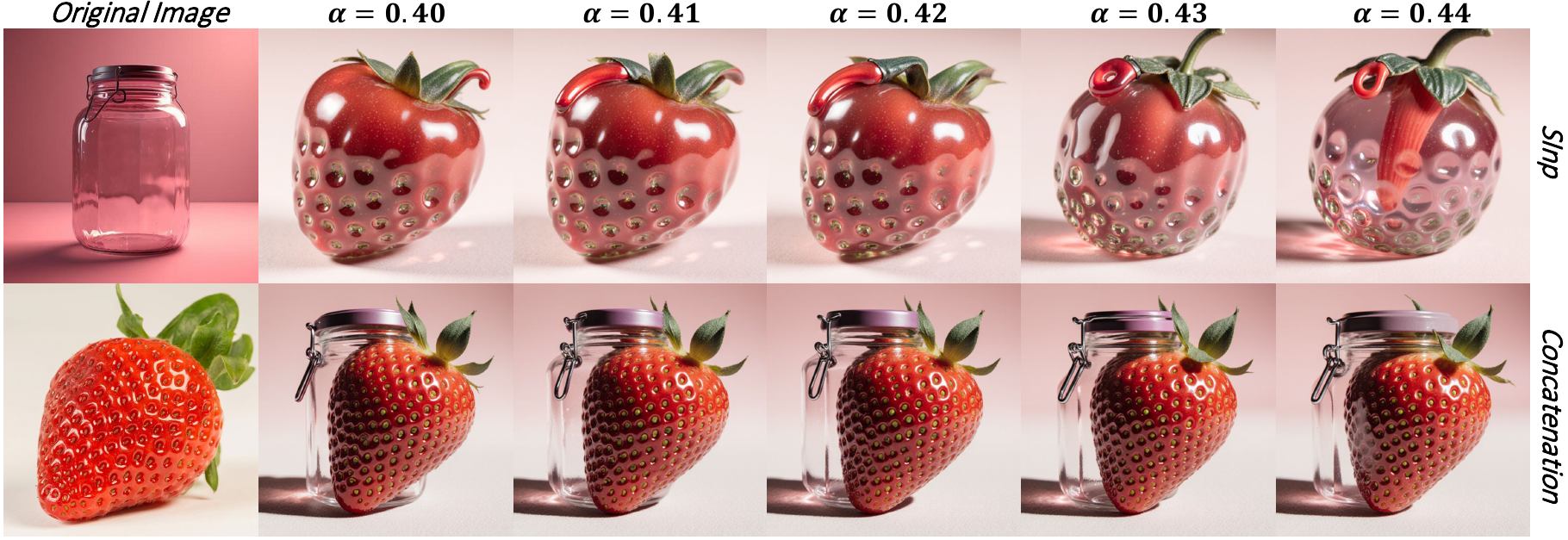}
  \vskip -0.15in
  \caption{Different MDeNoise generations across \(\alpha\).} % \textit{Concatenation (top)} produces disjointed outputs with visible input overlaps, while our \textit{SInP (bottom)} achieves unified and coherent blending.
  \label{fig:AB_concat_slerp}
   \vskip -0.15in
\end{wrapfigure}
%%%%%%%%%%%%%%%%%%%%%%%%%%%%%%%%%%%%%%%%%%%%%%%%%%%%%%%%%%%%%%%%%%%%%%%%%
\textit{\textbf{Discussion on MDeNoise: interpolate vs. concatenate.}} MDeNoise prioritizes fusing its two inputs, unlike BNoise which preserves them. While concatenation retains more input information, its rigid separation often creates disjointed representations and generations. However, interpolation enables seamless integration. To demonstrate this, we compare with a concatenation-fusion variant: \(z_{\text{SInp}}\) 
is replaced by \( z_{\text{SCat}}(\alpha) = \text{concat}(\alpha z_1, (1-\alpha) z_2)\) in \eqref{eq:MDeNoise} (Fig. \ref{fig:AB_concat_slerp}), which tends to produce isolated objects rather than a unified hybrid. Our interpolation instead creates a single, coherent entity with harmonious consistency.

\textbf{HSP:} Overall, for a given input \(D\), the hybrid sampling process combines the BNoise (\eqref{eq:BNoise}) and MDeNoise (\eqref{eq:MDeNoise}). To simplify the notation, we formalize this process as the function:
\begin{equation}
\label{eq:bsp}
I(\theta) = \operatorname{HSP}(D;\theta, \hat{\theta})=\mathcal{D}(x_0),
\end{equation}
where $\theta = \{\alpha, \beta_1, \beta_2, \epsilon\}$ are learnable parameters, and $\hat{\theta} = \{\gamma_{\text{den}}=5, \gamma_{\text{inv}}=0, \gamma_{\text{gen}}=4, T=999, t_{\text{den}}=652\}$ are fixed defaults in this paper.

\subsection{Efficient Adaptive Adjustment (EAA)}
\label{sec:EAA}
The HSP process yields distinct fusion results \( I(\theta)\) defined in \eqref{eq:bsp} with parameters \(\theta\), defaults \(\hat{\theta}\) and inputs \(D\), making parameter selection critical for high-quality synthesis. We propose an adaptive framework to jointly adjust $\theta=\{\alpha, \beta_1, \beta_2, \epsilon\}$, aiming to achieve both semantic coherence and visual fidelity. Inspired by prior work \cite{li2024tp2ocreativetextpairtoobject,xiong2024novel}, we first introduce a \textbf{Similarity Score (SS)} to guide this search: (\textit{For simplicity, input \(D\) and defaults \(\hat{\theta}\) are not shown.})
\begin{align}
S(\theta) = \underbrace{S_{I_1}(\theta) + S_{I_2}(\theta)}_{\text{visual similarity}} 
+ \underbrace{S_{T_1}(\theta) + S_{T_2}(\theta)}_{\text{semantic similarity}} - \underbrace{|S_{I_1}(\theta) - S_{I_2}(\theta)|}_{\text{visual balance}} 
- \underbrace{|S_{T_1}(\theta) - S_{T_2}(\theta)|}_{\text{semantic balance}},
\label{eq:score}
\end{align}
where \( S_{I_i}(\theta) \) ($i=1,2$) is the visual similarity between 
\( I(\theta) \) and the source image \( I_i \), computed via a DINO encoder \cite{oquab2024dinov2}, while \( S_{T_i}(\theta) \) ($i=1,2$)  is the semantic similarity between \( I(\theta) \) and the category label \( T_i \), measured using CLIP \cite{radford2021learning}.
This scoring function is designed to optimize two key objectives for successful fusion: (i) \emph{maximizing similarity}, and (ii) \emph{enforcing balance}. The first two terms ensure that the generated image \( I(\theta) \) retains high perceptual and semantic fidelity to both input images and their corresponding category labels. By maximizing similarity to both sources, these terms preserve the core features of the original concepts.
The final two terms—penalizing the absolute differences—explicitly enforce \textit{balance}, preventing the model from overfitting to one input and encouraging a fair integration of both objects’ features. Together, these components create a unified SS objective that balances fidelity and symmetry, offering a principled framework for optimizing feature fusion parameters.

\textbf{Our EAA Algorithm.}  
To maximize this objective \(S(\theta)\) in \eqref{eq:score}, we present a hierarchical adjustment strategy that learns the parameters \( \theta = \{\alpha, \beta_1, \beta_2, \epsilon\}\) using the acceptance threshold $Th=2.4$. The key loop iterates from $k=1$ to $K=3$, performing these steps:
\begin{itemize}
    \item[] \textcolor{red!70}{\textbf{\large\ding{192}}} \textbf{Sample (initial) Gaussian noise:} \(\epsilon\sim\mathcal{N}(0,I)\),  \textbf{initialize the parameters:} \(\alpha=0.5, \beta_1=\beta_2=1.0\).

    \item[] \textcolor{red!70}{\textbf{\large\ding{193}}}  \textbf{Searching $\alpha$:} Fixed \(\beta_1=\beta_2=1.0\) and \(\epsilon\), perform a golden section search \cite{teukolsky1992numerical} to find the optimal mixing factor $\alpha^*$: 
    \begin{align}
    \alpha^* = \arg\max_{\alpha \in [0, 1]} S(\alpha, \beta_1,\beta_2,\epsilon ).
    \label{eq:alpha_search}
    \end{align}
    \item[] \textcolor{red!70}{\textbf{\large\ding{194}}}  \textbf{Adjusting \(\beta_1, \beta_2\):} Fixed $\alpha^*,\epsilon$, 
    if \( S(\alpha^*, \beta_1, \beta_2,\epsilon) \leq Th \), then update the noise factors: 
    \begin{align}
    \left\{
    \begin{aligned}
    & \beta_1^*=\beta_1\ \& \  \beta_2^*=\arg\max_{\beta_2\in\mathbb{R}_+ } S(\alpha^*, \beta_1, \beta_2,\epsilon), & \text{if } S_1 > S_2, \\
    &  \beta_2^*=\beta_2\ \& \  \beta_1^*=\arg\max_{\beta_1\in\mathbb{R}_+} S(\alpha^*, \beta_1, \beta_2,\epsilon), & \text{otherwise}.
    \end{aligned}
    \right.,
    \label{eq:select_beta_updated}
    \end{align}
    where \(S_1 = S_{I_1} + S_{T_1} \), \(S_2 = S_{I_2} + S_{T_2}\), and $S_1>S_2$ indicates that the mixing noise favors the object $I_1$, and vice versa.
    \item[] \textbf{\large\ding{195}} \textbf{Acceptance criterion:} 
    \begin{align}
    \left\{
    \begin{aligned}
    & \epsilon^*=\epsilon\ \& \ \textbf{return} \ \theta^*=\{\alpha^*,\beta_1^*,\beta_2^*,\epsilon^*\}, & \text{if } S(\alpha^*,\beta_1^*,\beta_2^*,\epsilon) > Th, \\
    & \textbf{return} \ \theta^*=\{\alpha^*,\beta_1^*,\beta_2^*,\epsilon^*\}\ \& \ \textbf{break}, & \text{if } k> K, \\
    & \text{\textbf{turn to the step} \textcolor{red!70}{\textbf{\large\ding{192}}} \textbf{to resample}}\ \epsilon \ \& \ k++, & \text{otherwise}.
    \end{aligned}
    \right.,
    \label{eq:criterion}
    \end{align}
\end{itemize}
where the fused object image \(I(\theta) \) is defined in \eqref{eq:bsp}.
Our adaptive loop efficiently explores a low-dimensional yet expressive parameter space \( \theta = \{\alpha, \beta_1, \beta_2, \epsilon\} \), yielding conceptually balanced and perceptually smooth fusion results (Fig.~\ref{fig:ablation}). By reusing intermediate predictions and limiting optimization to scalar-level searches (via golden section search), the method enhances sample efficiency—avoiding the computational overhead of gradient-based latent-space backpropagation.

\textit{\textbf{Discussion on resampling \(\epsilon\).}} During our blending process, sampling random Gaussian noise can occasionally yield low-quality or failed fusions. While first-order optimization is an intuitive solution, it offers no significant advantage over simple zero-order resampling for diffusion generation, despite its higher cost \cite{ma2025inference}. Consequently, we adopt a zero-order resampling strategy to search for \(\epsilon\), and a small number of resamples \(K=3\) proves sufficient for high-quality fusion.
\textbf{\textit{For fair comparison, this resampling is disabled, \(K=1\), and the random seed is fixed at 42.}}

\section{Experiments}
\label{sec:experiment}

%%%%%%%%%%%%%%%%%%%%%%%%%%%%%%%%%%%%%%%%%%%%%%%%%%%%%%%%%%%%%%%%%%%%%%%%%%%%%

\subsection{Experimental Settings}
\label{subsec:exp_setting}
% These images include both synthetic and real examples.

\textbf{Datasets.} We introduce IIOF (Image-Image Object Fusion), a new benchmark of 780 image pairs derived from 40 objects across four classes (i.e., animals, fruits, artificial objects, and character figurines). Most images are from PIE-Bench~\cite{ju2023direct} and Pexels\footnote{\url{https://www.pexels.com/}}; figurines were self-captured for quality. To evaluate order-sensitive methods, we also generate all ordered pairs (1,560 total), ensuring a comprehensive and fair benchmark. \textit{\textbf{More details in Appdx.~\ref{sec:sup_Dataset}.}}

\begin{figure*}[t]
  \centering
  \includegraphics[width=0.97\linewidth]{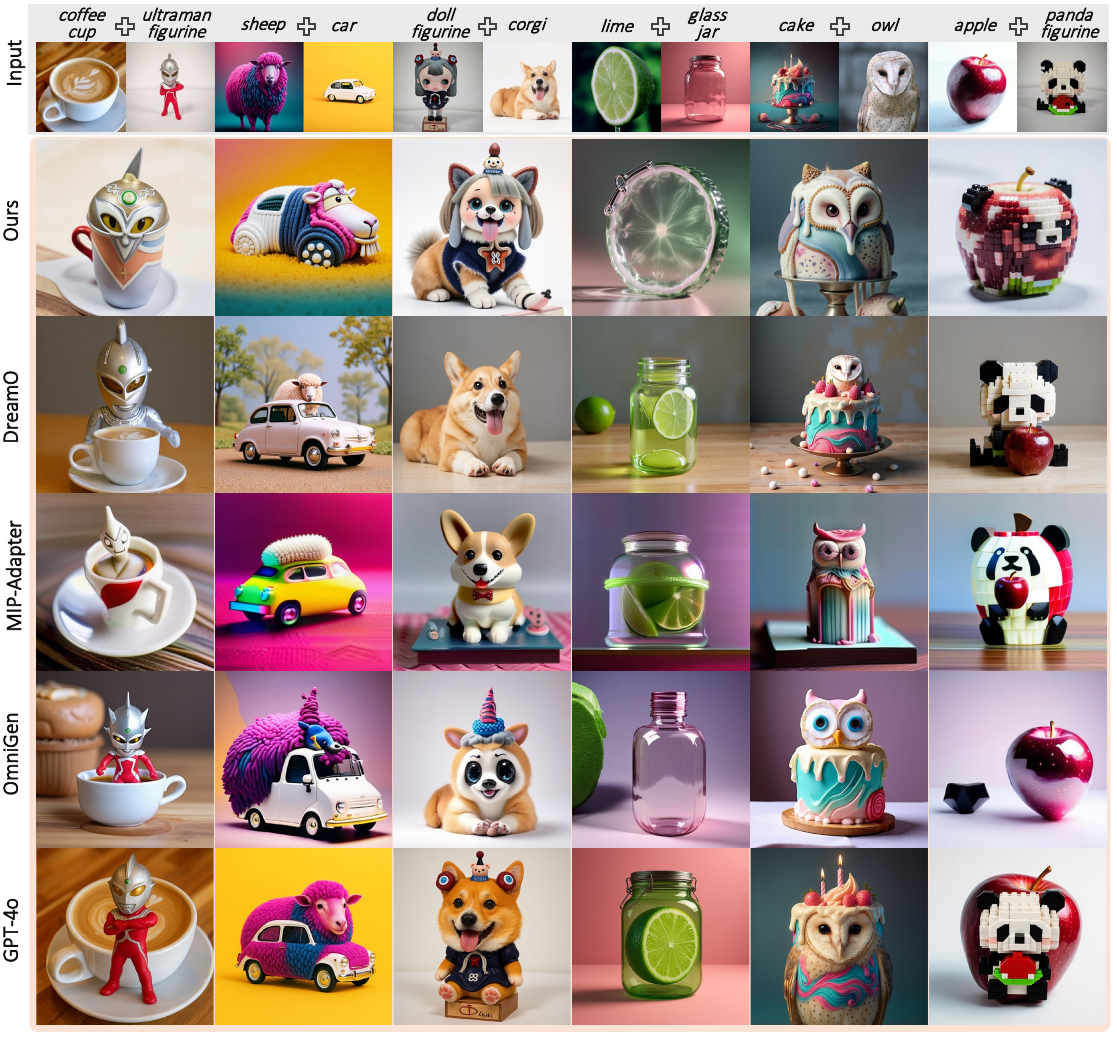}
  \vskip-0.15in
  \caption{ \textbf{Comparisons with Multi-Concept Generation Methods.} Our approach yields hybrid objects with improved structural coherence and visual balance over existing methods. }
  \label{fig:editedresults}
  \vskip -0.1in
\end{figure*}

\begin{figure*}[t]
  \centering
  \includegraphics[width=0.99\linewidth]{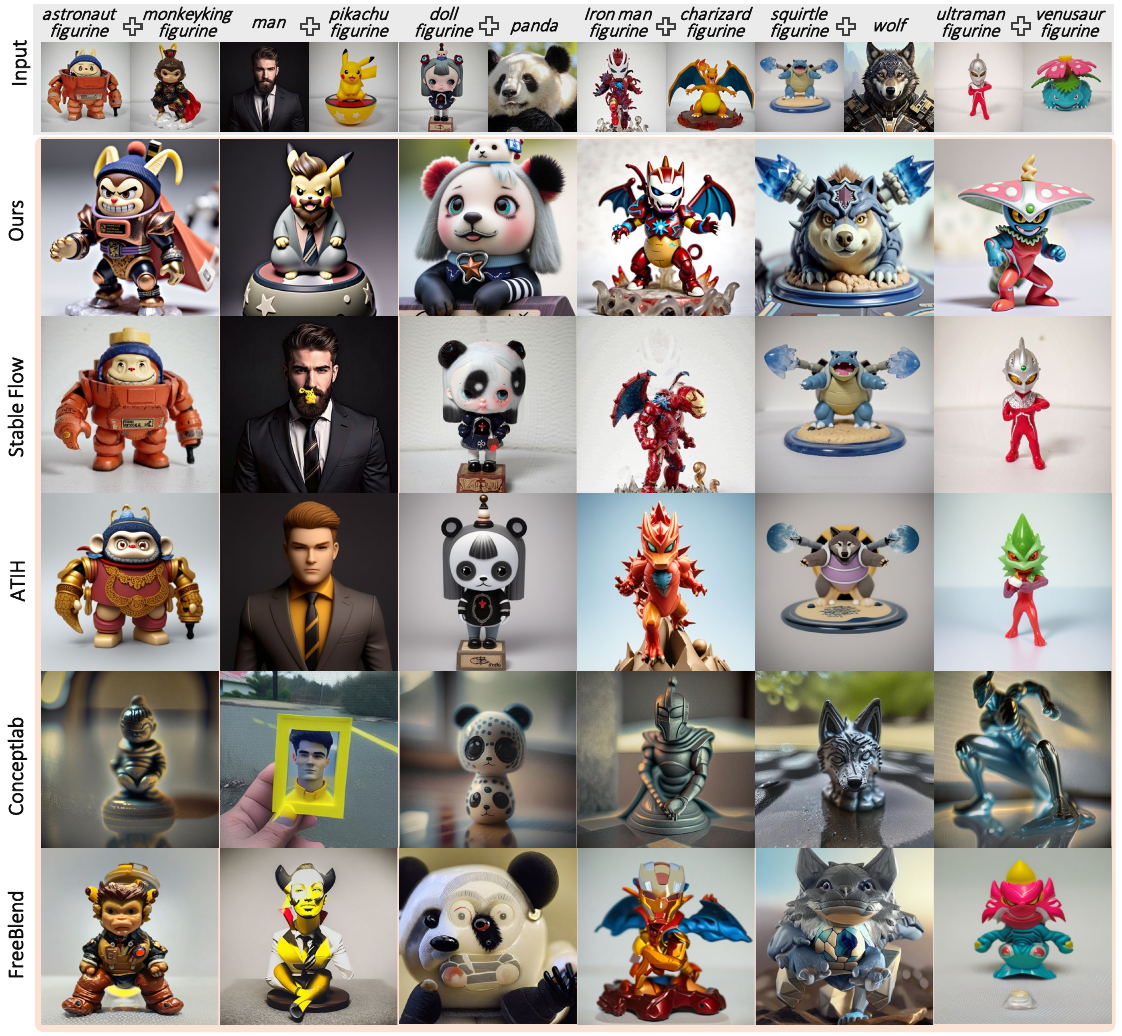}
  \vskip -0.1in
  \caption{\textbf{Comparisons with Mixing and Image Editing Methods.} Our method produces more coherent and balanced hybrids, while baselines often favor one concept or apply minimal edits.}
  \label{fig:mixresults}
  \vskip -0.1in
\end{figure*}

\textbf{Implementation Details.}  
Our method builds upon FLUX-Krea~\cite{flux1kreadev2025}, implementing 
\(\mathcal{E}_I\) with Redux~\cite{flux2024} for latent-space alignment. We generate all images at $512 \times 512$ resolution using the FlowMatchEulerDiscreteScheduler~\cite{lipmanflow} with 20 denoising steps.
For the Efficient Adaptive Adjustment (EAA) module, we use Grounded-SAM~\cite{ren2024grounded} and the query \textit{“most prominent object”} to localize main regions for visual and semantic similarity computation. For each parameter ($\alpha$ and $\beta$), the search is limited to at most 10 image generations, respectively. All experiments are conducted on two NVIDIA RTX 4090 GPUs.

\textbf{Evaluation Metrics.}  
To evaluate our method, we use two metric families: Semantic Alignment (SA) and Single-entity Coherence (SCE).
SA is computed on the generated prompt $P_G$ using VQAScore~\cite{lin2024evaluating} and LLaVA\mbox{-}Critic~\cite{xiong2025llava}. VQAScore employs CLIP\mbox{-}FlanT5~\cite{roberts2022t5x} and LLaVA~\cite{liu2023visual}, denoted as $\mathrm{VQA}^{\mathrm{SA}}_{\mathrm{T5}}$ and $\mathrm{VQA}^{\mathrm{SA}}_{\mathrm{LLaVA}}$, respectively; the LLaVA\mbox{-}Critic score is $\mathrm{LC}^{\mathrm{SA}}$.
SCE assesses if the image forms a unified concept by asking: \textit{``A photo of a seamless fusion of \textless$T_1$\textgreater{} and \textless$T_2$\textgreater{} into a single coherent entity.''} Its scores are $\mathrm{VQA}^{\mathrm{SCE}}_{\mathrm{T5}}$, $\mathrm{VQA}^{\mathrm{SCE}}_{\mathrm{LLaVA}}$, and $\mathrm{LC}^{\mathrm{SCE}}$.
We also compute the SS score and the balance metric $B_{\text{sim}} = |S_{I_1}(\theta) - S_{I_2}(\theta)| + |S_{T_1}(\theta) - S_{T_2}(\theta)|$, where $S_{T_i}(\theta)$ are normalized to $[0,1]$ using empirical bounds 0.15 and 0.45 to align the scales of visual and textual modalities.

\subsection{Main Results}
\label{subsec:main_res}

We compare with leading methods across three categories: (i) multi-concept generation (e.g., OmniGen~\cite{xiao2024omnigen}, FreeCustom~\cite{ding2024freecustom}, MIP-Adapter~\cite{huang2025resolving}, DreamO~\cite{mou2025dreamo}), (ii) mixing-based (e.g., ATIH~\cite{xiong2024novel}, Conceptlab~\cite{Richardson2024conceptlab}, FreeBlend~\cite{zhou2025freeblend}), and (iii) image editing (e.g., Stable Flow~\cite{avrahami2024stable}). We also include qualitative results from GPT-4o~\cite{openai2023chatgpt}. Inputs vary: multi-concept methods use two images and a text prompt; ATIH and Stable Flow use one image and text; Conceptlab uses text only. \textit{\textbf{More examples in Appdx.~\ref{sec:more_results}.}}

%%%%%%%%%%%%%%%%%%%%%%%%%%%%%%%%%%%%%%%%%%%%%%%%%%%%%%%%%%%%%%%%%%%%%%%%%%%%%%

\textbf{Qualitative Comparison.}
Fig.~\ref{fig:editedresults} compares our method with multi-concept generation baselines (e.g., MIP-Adapter, OmniGen, DreamO, GPT-4o), highlighting two observations. First, baselines output often merely overlay features rather than fusing them—for example, \textit{a lime enclosed in a glass jar without integration}—while our method creates a coherent hybrid. Second, baselines frequently favor one concept, such as generating either a doll or a corgi but not a unified blend. In contrast, our approach balances both concepts, producing structurally unified and semantically consistent results. This demonstrates our method’s superior ability to achieve fine-grained visual fusion.

%%%%%%%%%%%%%%%%%%%%%%%%%%%%%%%%%%%%%%%%%%%%%%%%%%%%%%%%%%%%%%%%%%%%%%%%%%%%%%
\begin{table}[t]
\centering
\setlength{\tabcolsep}{7pt}
\renewcommand{\arraystretch}{1.1}
\caption{Quantitative comparisons on our IIOF dataset.}
\label{tab:quant_cmp}
%\vskip -0.1in
\resizebox{0.97\linewidth}{!}{
\begin{tabular}{c||c|c|c|c|c|c|c|c}
\toprule[1.2pt]
\addlinespace[1pt]  
Models & $\mathrm{VQA}^{\mathrm{SA}}_{\mathrm{T5}}$$\uparrow$      & $\mathrm{VQA}^{\mathrm{SCE}}_{\mathrm{T5}}$$\uparrow$     & $\mathrm{LC}^{\mathrm{SA}}$ $\uparrow$  &$\mathrm{LC}^{\mathrm{SCE}}$$\uparrow$   &  $\mathrm{VQA}^{\mathrm{SA}}_{\mathrm{LLaVA}}\uparrow$  & $\mathrm{VQA}^{\mathrm{SCE}}_{\mathrm{LLaVA}}$$\uparrow$ &$SS\uparrow$  & $B$sim$\downarrow$\\ \hline
\rowcolor{green!10} \textbf{Our VMDiff}           &\textcolor{red!70}{0.639}  & \textcolor{red!70}{0.540}  & \textcolor{red!70}{8.372 } & 
\textcolor{red!70}{8.392 } &
\textcolor{red!70}{0.390}      & \textcolor{blue!70}{0.413} & \textcolor{red!70}{2.068} & \textcolor{red!70}{0.324}\\
FreeCustom \small{(CVPR \cite{ding2024freecustom})}        & 0.579  & 0.452  & 6.958 & 6.946 & 0.360      & 0.388  & 1.580 & 0.776 \\
MIP-Adapter \small{(AAAI \cite{huang2025resolving})} & \textcolor{blue!70}{0.621} &  \textcolor{blue!70}{0.512}   & \textcolor{blue!70}{8.301}  & \textcolor{blue!70}{8.076} & \textcolor{blue!70}{0.389}   & \textcolor{red!70}{0.417}  &1.866 &0.483 \\
OmniGen \small{(CVPR \cite{xiao2024omnigen})} & 0.570 &  0.469   & 7.550  & 7.233 & 0.352 & 0.348  &1.705 
&0.617 \\
Conceptlab \small{(TOG \cite{Richardson2024conceptlab})}       & 0.573  & 0.483 & 7.589 & 7.728 & 0.362  & 0.395    &-- & --\\
ATIH \small{(NeurIPS \cite{xiong2024novel} )}        & 0.523  & 0.465 & 7.275 & 6.816 &0.317 &0.367   & -- & -- \\
Stable Flow \small{(CVPR \cite{avrahami2024stable})}        & 0.460   & 0.372 & 6.020 & 5.024 & 0.266  & 0.294    &-- & --\\
DreamO \small{(SIGGRAPH Asia \cite{mou2025dreamo} )}       & 0.591   & 0.467 & 7.592 & 7.013 & 0.370  & 0.346   &1.793 & 0.644\\
FreeBlend \small{(arXiv \cite{zhou2025freeblend})}  & 0.588   & 0.507 & 7.836 & 7.788 & 0.341  & 0.383    &\textcolor{blue!70}{1.870} & \textcolor{blue!70}{0.479}\\
\bottomrule
\end{tabular}}
\vskip -0.15in
\end{table}
%%%%%%%%%%%%%%%%%%%%%%%%%%%%%%%%%%%%%%%%%%%%%%%%%%%%%%%%%%%%%%%%%%%%%%%%%%%%%% 

Fig.~\ref{fig:mixresults} qualitatively compares our method with mixing/editing baselines (e.g., Conceptlab, ATIH, FreeBlend, Stable Flow). Conceptlab often biases toward one concept, while Stable Flow and ATIH make only subtle edits, such as color or texture transfer. FreeBlend frequently loses original information and yields fragmented outputs. In contrast, our approach synthesizes novel objects that structurally and visually integrate both concepts, achieving a deeper, more harmonious fusion and demonstrating superior blending capability.

%%%%%%%%%%%%%%%%%%%%%%%%%%%%%%%%%%%%%%%%%%%%%%%%%%%%%%%%%%%%%%%%%%%%%%%%%%%%%% 
 \textbf{Quantitative Comparison.}  
Table~\ref{tab:quant_cmp} presents quantitative comparisons on key metrics,  including $\mathrm{VQA}^{\mathrm{SA}}_{\mathrm{T5}}$, $\mathrm{VQA}^{\mathrm{SA}}_{\mathrm{LLaVA}}$, $\mathrm{VQA}^{\mathrm{SCE}}_{\mathrm{T5}}$, ,$\mathrm{VQA}^{\mathrm{SCE}}_{\mathrm{LLaVA}}$, $\mathrm{LC}^{\mathrm{SA}}$, $\mathrm{LC}^{\mathrm{SCE}}$,  similarity score (SS), and fusion balance $B_{\text{sim}}$. Although MIP attains the highest $\mathrm{VQA}^{\mathrm{SCE}}_{\mathrm{LLaVA}}$, it ranks only second or below on the other VQA, LC, SS, and $B_{\text{sim}}$ metrics, indicating that its improvements are not holistic. In contrast, our method consistently outperforms all baselines on most metrics, demonstrating strong capability in generating coherent and natural blended objects.
These results reinforce our qualitative findings and confirm the effectiveness of our approach in achieving high-quality visual fusion.

\begin{wrapfigure}{r}{0.5\textwidth}
\vskip -0in
    \centering
    \includegraphics[width=1\linewidth]{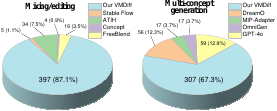}
    \vskip -0.1in
    \caption{User studies.}  %Our approach received the highest vote share in both groups.  comparing our VMDiff with mixing/editing (left) and multi-concept generation (right)
    \label{fig:user_study} 
    \vskip -0.1in
\end{wrapfigure}
\textbf{User Study.} 
To evaluate the perceptual quality of our fusions, we conducted two user studies (Fig. \ref{fig:user_study}). 76 participants each rated 12 results—6 from \textit{Multi-Concept Generation} and 6 from \textit{Mixing/Editing}—yielding 912 total votes. Our VMDiff received the highest preference in both groups: \textbf{67.3\%} and \textbf{87.1\%}, respectively. GPT-4o and ATIH ranked second, but with significantly lower votes (12.9\% and 7.5\%). These results indicate that our VMDiff aligns better with human preferences in visual coherence and creativity. \textit{\textbf{More details in Appdx.~\ref{sec:sup_user_stu}.}}

%%%%%%%%%%%%%%%%%%%%%%%%%%%%%%%%%%%%%%%%%%%%%%%%%%%%%%%%%%%%%%%%%%%%%%%%%%%%%%
\begin{table}[t]
\centering
\setlength{\tabcolsep}{7pt}
\renewcommand{\arraystretch}{1.1}
\caption{Quantitative ablation study on our IIOF dataset.}
\label{tab:quant_module_ablaution}
%\vskip -0.1in
\resizebox{0.95\linewidth}{!}{
\begin{tabular}{c||c|c|c|c|c|c|c|c}
\toprule[1.2pt]
\addlinespace[1pt]  
Models & $\mathrm{VQA}^{\mathrm{SA}}_{\mathrm{T5}}$$\uparrow$      & $\mathrm{VQA}^{\mathrm{SCE}}_{\mathrm{T5}}$$\uparrow$     & $\mathrm{LC}^{\mathrm{SA}}$ $\uparrow$  &$\mathrm{LC}^{\mathrm{SCE}}$$\uparrow$   &  $\mathrm{VQA}^{\mathrm{SA}}_{\mathrm{LLaVA}}\uparrow$  & $\mathrm{VQA}^{\mathrm{SCE}}_{\mathrm{LLaVA}}$$\uparrow$ &$SS\uparrow$  & $B$sim$\downarrow$\\ \hline
 Baseline 1           &0.497  & 0.438  & 7.261 & 
7.077 &
0.287     & 0.314 & 1.570 & 0.682\\
Baseline 2        & 0.508 & 0.441  & 7.426 & 7.291 & 0.298      & 0.325  & 1.586 & 0.693 \\
Baseline 2+$\alpha$-search  & 0.625 &  0.532   & 8.278  & 8.276 & 0.382   & 0.405  &2.025  &0.358 \\
Baseline 2+$\alpha$-search+$\beta_1,\beta_2$-search & \textbf{0.639} &  \textbf{0.540}   & \textbf{8.372}  & \textbf{8.392} & \textbf{0.390} & \textbf{0.413}  & \textbf{2.068} 
&\textbf{0.324} \\

\bottomrule
\end{tabular}}
\vskip -0.1in
\end{table}
%%%%%%%%%%%%%%%%%%%%%%%%%%%%%%%%%%%%%%%%%%%%%%%%%%%%%%%%%%%%%%%%%%%%%%%%%%%%%% 
\subsection{Ablation Study }
\label{subsec:parameter}
\begin{wrapfigure}{r}{0.65\textwidth}
\vskip -0.1in
   \centering
   \includegraphics[width=0.98\linewidth]{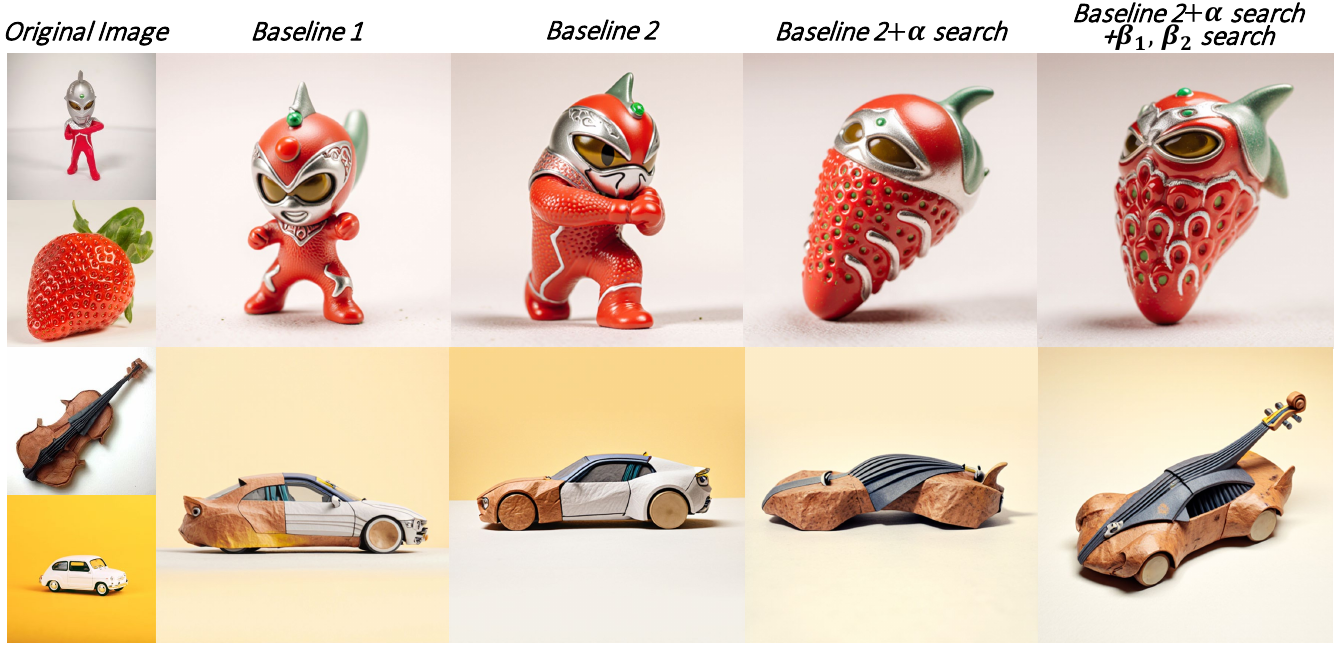}
   \vskip -0.1in
   \caption{\textbf{Ablation study in VMDiff.} \textit{Noise refinement} improves detail and structure, while \textit{adaptive $\alpha$ and $\beta$ search} progressively enhance semantic balance and visual coherence.}
   \label{fig:ablation}
   \vskip -0.05in
\end{wrapfigure}

\begin{figure*}[t]
    \centering
    \includegraphics[width=0.97\linewidth]{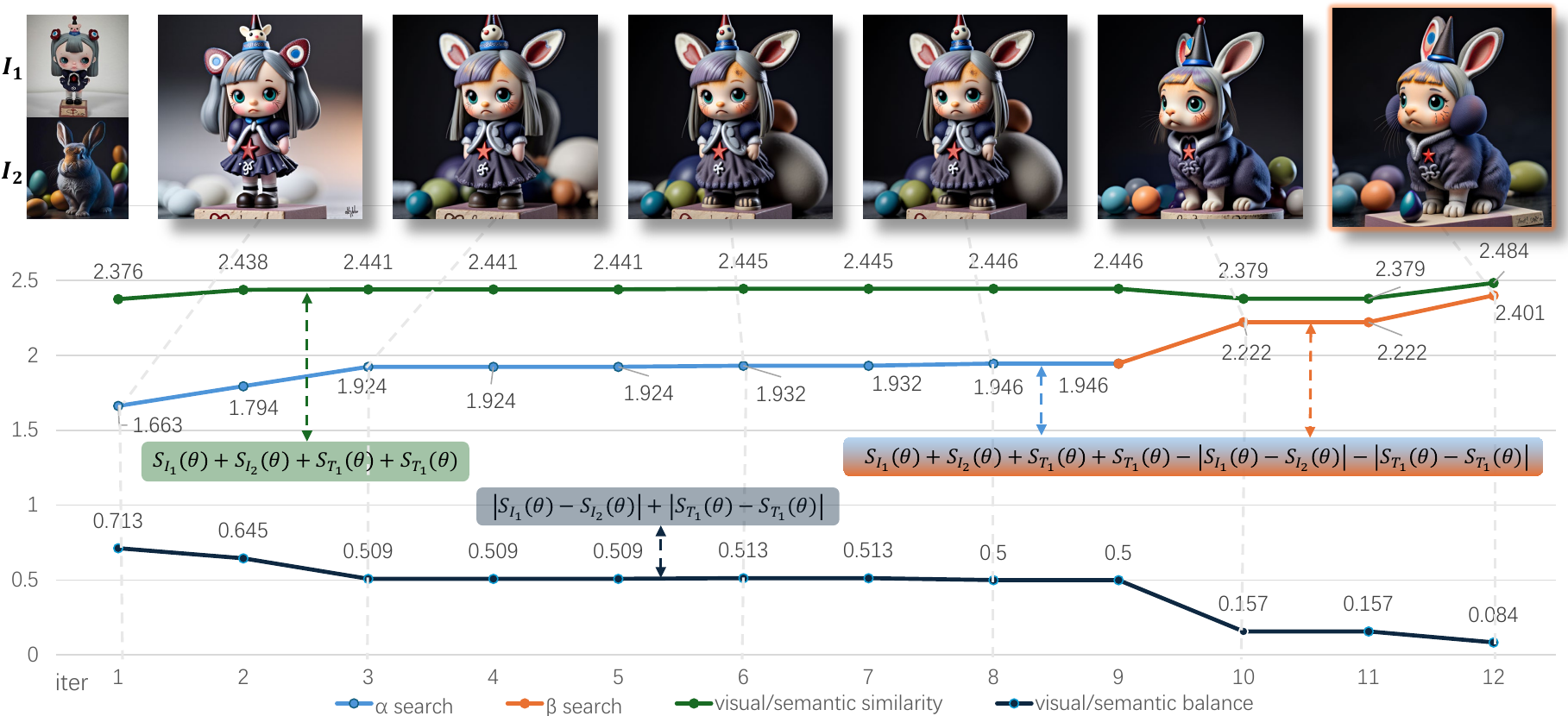}
    \vskip -0.1in
    \caption{
\textbf{Visualizing the updated process of our EAA} based on two input images $I_1$ (\textit{doll figurine}) and $I_2$ (\textit{rabbit}). The $\alpha$ parameter (blue) improves fusion quality, while $\beta$ (orange) enhances semantic balance. The green curve (similarity) rises and the dark blue curve (imbalance) falls over iterations. The final output is a coherent hybrid with high similarity and minimal imbalance.}
\label{fig:iterativeprocess}
\vskip -0.15in
\end{figure*}

We conducted an ablation study to evaluate the contributions of our VMDiff's key components, as shown in Fig.~\ref{fig:ablation} and Table~\ref{tab:quant_module_ablaution}. Progressively adding each element—\textbf{(i)} \textit{baseline 1:} random noise+MDeNoise (\(\alpha=0.5\)), \textbf{(ii)} \textit{baseline 2:} baseline 1+BNoise (\(\beta_1=\beta_2=1\)), \textbf{(iii)} baseline 2 + MDeNoise ($\alpha$ search), and \textbf{(iv)} baseline 2 + BNoise (\(\beta_1,\beta_2\) search) + MDeNoise ($\alpha$ search)—yielded consistent improvements.
Without noise refinement, outputs lacked detail. Its 
%%%%%%%%%%%%%%%%%%%%%%%%%%%%%%%%%%%%%%%%%%%%%%%%%%%%%%%%%%%%%%%%%%%%%%%%%%%%%% 
%%%%%%%%%%%%%%%%%%%%%%%%%%%%%%%%%%%%%%%%%%%%%%%%%%%%%%%%%%%%%%%%%%%%%%%%%%%%%% 
inclusion enhanced structural fidelity and preserved input features. Adaptive $\alpha$ improved fusion balance, while adaptive $\beta$ refined noise influence for greater visual harmony.
Fig.~\ref{fig:iterativeprocess} illustrates the optimization process for a representative case (\textit{doll figurine + rabbit}). Throughout iterations, similarity $S(\theta)$ (green) increased steadily, while the blending balance metric (dark blue) decreased. The $\alpha$ search (light blue) rapidly boosted similarity, and $\beta$ search (orange) smoothed visual-textual alignment. These results confirm that our EAA design effectively optimizes both similarity and symmetry for high-quality blending. \textit{\textbf{Limitations are discussed in Appdx.~\ref{sec:limitation}.}}

\section{Conclusion}
In this paper, we presented VMDiff, a novel unified and controllable framework for visual concept fusion that synthesizes coherent new objects directly from two input images. Our approach enables fine-grained control by semantically integrating concepts at both the noise and latent levels. VMDiff consists of two core components: (1) a hybrid sampling process that constructs optimized semantic noise through guided denoising and inversion, followed by a curvature-aware latent fusion using spherical interpolation, and (2) an efficient adaptive adjustment algorithm that refines fusion parameters via a lightweight, score-driven search. Experimental results on a curated benchmark demonstrate VMDiff's superior performance, excelling in semantic consistency, visual harmony, and user-rated creativity, thereby establishing a new paradigm for hybrid object synthesis. This work offers practical and valuable insights for professionals developing combinational characters, directly applicable to diverse fields from film and animation to figures and industrial design.

\bibliography{iclr2026_conference}

\begin{thebibliography}{53}
\providecommand{\natexlab}[1]{#1}
\providecommand{\url}[1]{\texttt{#1}}
\expandafter\ifx\csname urlstyle\endcsname\relax
  \providecommand{\doi}[1]{doi: #1}\else
  \providecommand{\doi}{doi: \begingroup \urlstyle{rm}\Url}\fi

\bibitem[Albergo \& Vanden-Eijnden(2023)Albergo and Vanden-Eijnden]{albergo2022building}
Michael~S Albergo and Eric Vanden-Eijnden.
\newblock Building normalizing flows with stochastic interpolants.
\newblock In \emph{Proceedings of the International Conference on Learning Representations (ICLR)}, 2023.

\bibitem[Avrahami et~al.(2025)Avrahami, Patashnik, Fried, Nemchinov, Aberman, Lischinski, and Cohen-Or]{avrahami2024stable}
Omri Avrahami, Or~Patashnik, Ohad Fried, Egor Nemchinov, Kfir Aberman, Dani Lischinski, and Daniel Cohen-Or.
\newblock Stable flow: Vital layers for training-free image editing.
\newblock In \emph{Proceedings of the IEEE/CVF conference on Computer Vision and Pattern Recognition (CVPR)}, pp.\  7877--7888, 2025.

\bibitem[Bai et~al.(2025)Bai, Shao, Zhou, Qi, Xu, Xiong, and Xie]{bai2024zigzag}
Lichen Bai, Shitong Shao, Zikai Zhou, Zipeng Qi, Zhiqiang Xu, Haoyi Xiong, and Zeke Xie.
\newblock Zigzag diffusion sampling: Diffusion models can self-improve via self-reflection.
\newblock In \emph{Proceedings of the International Conference on Learning Representations (ICLR)}, 2025.

\bibitem[{Black Forest Labs}(2024)]{flux2024}
{Black Forest Labs}.
\newblock {Flux}.
\newblock \url{https://github.com/black-forest-labs/flux}, 2024.
\newblock Accessed: 2025-05-07.

\bibitem[Boden(2004)]{boden2004creative}
Margaret~A Boden.
\newblock \emph{The creative mind: Myths and mechanisms}.
\newblock Routledge, 2004.

\bibitem[Brooks et~al.(2023)Brooks, Holynski, and Efros]{brooks2023instructpix2pix}
Tim Brooks, Aleksander Holynski, and Alexei~A. Efros.
\newblock Instructpix2pix: Learning to follow image editing instructions.
\newblock In \emph{Proceedings of the IEEE/CVF Conference on Computer Vision and Pattern Recognition (CVPR)}, pp.\  18392--18402, 2023.

\bibitem[Ceylan et~al.(2023)Ceylan, Huang, and Mitra]{ceylan2023pix2video}
Duygu Ceylan, Chun-Hao~P Huang, and Niloy~J Mitra.
\newblock Pix2video: Video editing using image diffusion.
\newblock In \emph{Proceedings of the IEEE/CVF International Conference on Computer Vision (ICCV)}, pp.\  23206--23217, 2023.

\bibitem[Chen et~al.(2024)Chen, Chen, Zhang, Wang, Yang, Wang, Cai, Yang, Liu, and Lin]{chen2024gaussianeditor}
Yiwen Chen, Zilong Chen, Chi Zhang, Feng Wang, Xiaofeng Yang, Yikai Wang, Zhongang Cai, Lei Yang, Huaping Liu, and Guosheng Lin.
\newblock Gaussianeditor: Swift and controllable 3d editing with gaussian splatting.
\newblock In \emph{Proceedings of the IEEE/CVF conference on Computer Vision and Pattern Recognition (CVPR)}, pp.\  21476--21485, 2024.

\bibitem[Ding et~al.(2024)Ding, Zhao, Wang, Yang, Liu, Chen, and Shen]{ding2024freecustom}
Ganggui Ding, Canyu Zhao, Wen Wang, Zhen Yang, Zide Liu, Hao Chen, and Chunhua Shen.
\newblock Freecustom: Tuning-free customized image generation for multi-concept composition.
\newblock In \emph{Proceedings of the IEEE/CVF Conference on Computer Vision and Pattern Recognition (CVPR)}, pp.\  9089--9098, 2024.

\bibitem[Dong \& Han(2023)Dong and Han]{Dong2023prompt}
Xiaoyue Dong and Shumin Han.
\newblock Prompt tuning inversion for text-driven image editing using diffusion models.
\newblock In \emph{Proceedings of the IEEE/CVF International Conference on Computer Vision (ICCV)}, pp.\  7430--7440, 2023.

\bibitem[Gal et~al.(2023)Gal, Alaluf, Atzmon, Patashnik, Bermano, Chechik, and Cohen-Or]{Gal2023personalizeT2I}
Rinon Gal, Yuval Alaluf, Yuval Atzmon, Or~Patashnik, Amit~Haim Bermano, Gal Chechik, and Daniel Cohen-Or.
\newblock An image is worth one word: Personalizing text-to-image generation using textual inversion.
\newblock In \emph{Proceedings of the International Conference on Learning Representations (ICLR)}, 2023.

\bibitem[Gu et~al.(2023)Gu, Wang, Wu, Shi, Chen, Fan, Xiao, Zhao, Chang, Wu, et~al.]{gu2023mix}
Yuchao Gu, Xintao Wang, Jay~Zhangjie Wu, Yujun Shi, Yunpeng Chen, Zihan Fan, Wuyou Xiao, Rui Zhao, Shuning Chang, Weijia Wu, et~al.
\newblock Mix-of-show: Decentralized low-rank adaptation for multi-concept customization of diffusion models.
\newblock \emph{Proceedings of the Advances in Neural Information Processing Systems (NeurIPS)}, 36:\penalty0 15890--15902, 2023.

\bibitem[Han et~al.(2023)Han, Li, Zhang, Milanfar, Metaxas, and Yang]{han2023svdiff}
Ligong Han, Yinxiao Li, Han Zhang, Peyman Milanfar, Dimitris Metaxas, and Feng Yang.
\newblock Svdiff: Compact parameter space for diffusion fine-tuning.
\newblock In \emph{Proceedings of the IEEE/CVF International Conference on Computer Vision (ICCV)}, pp.\  7323--7334, 2023.

\bibitem[Haque et~al.(2023)Haque, Tancik, Efros, Holynski, and Kanazawa]{haque2023instruct}
Ayaan Haque, Matthew Tancik, Alexei~A Efros, Aleksander Holynski, and Angjoo Kanazawa.
\newblock Instruct-nerf2nerf: Editing 3d scenes with instructions.
\newblock In \emph{Proceedings of the IEEE/CVF International Conference on Computer Vision (ICCV)}, pp.\  19740--19750, 2023.

\bibitem[Huang et~al.(2025)Huang, Fu, Liu, Jiang, Yu, and Song]{huang2025resolving}
Qihan Huang, Siming Fu, Jinlong Liu, Hao Jiang, Yipeng Yu, and Jie Song.
\newblock Resolving multi-condition confusion for finetuning-free personalized image generation.
\newblock In \emph{Proceedings of the AAAI Conference on Artificial Intelligence (AAAI)}, pp.\  3707--3714, 2025.

\bibitem[Ju et~al.(2024)Ju, Zeng, Bian, Liu, and Xu]{ju2023direct}
Xuan Ju, Ailing Zeng, Yuxuan Bian, Shaoteng Liu, and Qiang Xu.
\newblock Direct inversion: Boosting diffusion-based editing with 3 lines of code.
\newblock In \emph{Proceedings of the International Conference on Learning Representations (ICLR)}, 2024.

\bibitem[Ke et~al.(2023)Ke, Liu, Zhu, Zhao, and Lau]{ke2023neural}
Zhanghan Ke, Yuhao Liu, Lei Zhu, Nanxuan Zhao, and Rynson~WH Lau.
\newblock Neural preset for color style transfer.
\newblock In \emph{Proceedings of the IEEE/CVF conference on Computer Vision and Pattern Recognition (ICCV)}, pp.\  14173--14182, 2023.

\bibitem[Kumari et~al.(2023)Kumari, Zhang, Zhang, Shechtman, and Zhu]{kumari2023multi}
Nupur Kumari, Bingliang Zhang, Richard Zhang, Eli Shechtman, and Jun-Yan Zhu.
\newblock Multi-concept customization of text-to-image diffusion.
\newblock In \emph{Proceedings of the IEEE/CVF Conference on Computer Vision and Pattern Recognition (CVPR)}, pp.\  1931--1941, 2023.

\bibitem[Lee et~al.(2025)Lee, Ebbecke, Millon, Beddow, Zhuo, García-Ferrero, Esparraguera, Petrescu, Saß, Menezes, and Perez]{flux1kreadev2025}
Sangwu Lee, Titus Ebbecke, Erwann Millon, Will Beddow, Le~Zhuo, Iker García-Ferrero, Liam Esparraguera, Mihai Petrescu, Gian Saß, Gabriel Menezes, and Victor Perez.
\newblock {FLUX.1 Krea [dev]}.
\newblock \url{https://github.com/krea-ai/flux-krea}, 2025.

\bibitem[Li et~al.(2024)Li, Zhang, and Yang]{li2024tp2ocreativetextpairtoobject}
Jun Li, Zedong Zhang, and Jian Yang.
\newblock Tp2o: Creative text pair-to-object generation using balance swap-sampling.
\newblock In \emph{Proceedings of the European Conference on Computer Vision (ECCV)}, pp.\  92--111, 2024.

\bibitem[Liew et~al.(2022)Liew, Yan, Zhou, and Feng]{Liew2022Magicmix}
Jun~Hao Liew, Hanshu Yan, Daquan Zhou, and Jiashi Feng.
\newblock Magicmix: Semantic mixing with diffusion models.
\newblock \emph{arXiv preprint arXiv:2210.16056}, 2022.

\bibitem[Lin et~al.(2024)Lin, Pathak, Li, Li, Xia, Neubig, Zhang, and Ramanan]{lin2024evaluating}
Zhiqiu Lin, Deepak Pathak, Baiqi Li, Jiayao Li, Xide Xia, Graham Neubig, Pengchuan Zhang, and Deva Ramanan.
\newblock Evaluating text-to-visual generation with image-to-text generation.
\newblock In \emph{Proceedings of the European Conference on Computer Vision (ECCV)}, pp.\  366--384, 2024.

\bibitem[Lipman et~al.(2022)Lipman, Chen, Ben-Hamu, Nickel, and Le]{lipmanflow}
Yaron Lipman, Ricky~TQ Chen, Heli Ben-Hamu, Maximilian Nickel, and Matthew Le.
\newblock Flow matching for generative modeling.
\newblock In \emph{Proceedings of the International Conference on Learning Representations (ICLR)}, 2022.

\bibitem[Liu et~al.(2023{\natexlab{a}})Liu, Li, Wu, and Lee]{liu2023visual}
Haotian Liu, Chunyuan Li, Qingyang Wu, and Yong~Jae Lee.
\newblock Visual instruction tuning.
\newblock In \emph{Proceedings of the Advances in Neural Information Processing Systems (NeurIPS)}, pp.\  34892--34916, 2023{\natexlab{a}}.

\bibitem[Liu et~al.(2021)Liu, Li, Du, Tenenbaum, and Torralba]{liu2021learning}
Nan Liu, Shuang Li, Yilun Du, Josh Tenenbaum, and Antonio Torralba.
\newblock Learning to compose visual relations.
\newblock In \emph{Proceedings of the Advances in Neural Information Processing Systems (NeurIPS)}, pp.\  23166--23178, 2021.

\bibitem[Liu et~al.(2022)Liu, Li, Du, Torralba, and Tenenbaum]{liu2022compositional}
Nan Liu, Shuang Li, Yilun Du, Antonio Torralba, and Joshua~B Tenenbaum.
\newblock Compositional visual generation with composable diffusion models.
\newblock In \emph{Proceedings of the European Conference on Computer Vision (ECCV)}, pp.\  423--439, 2022.

\bibitem[Liu et~al.(2024)Liu, Zhang, Li, Lin, and Jia]{liu2024video}
Shaoteng Liu, Yuechen Zhang, Wenbo Li, Zhe Lin, and Jiaya Jia.
\newblock Video-p2p: Video editing with cross-attention control.
\newblock In \emph{Proceedings of the IEEE/CVF Conference on Computer Vision and Pattern Recognition (CVPR)}, pp.\  8599--8608, 2024.

\bibitem[Liu et~al.(2023{\natexlab{b}})Liu, Zhang, Shen, Zheng, Zhu, Feng, Liu, Zhao, Zhou, and Cao]{liu2023cones}
Zhiheng Liu, Yifei Zhang, Yujun Shen, Kecheng Zheng, Kai Zhu, Ruili Feng, Yu~Liu, Deli Zhao, Jingren Zhou, and Yang Cao.
\newblock Cones 2: customizable image synthesis with multiple subjects.
\newblock In \emph{Proceedings of the Advances in Neural Information Processing Systems (NeurIPS)}, pp.\  57500--57519, 2023{\natexlab{b}}.

\bibitem[Ma et~al.(2025)Ma, Tong, Jia, Hu, Su, Zhang, Yang, Li, Jaakkola, Jia, et~al.]{ma2025inference}
Nanye Ma, Shangyuan Tong, Haolin Jia, Hexiang Hu, Yu-Chuan Su, Mingda Zhang, Xuan Yang, Yandong Li, Tommi Jaakkola, Xuhui Jia, et~al.
\newblock Scaling inference time compute for diffusion models.
\newblock In \emph{Proceedings of the IEEE/CVF Conference on Computer Vision and Pattern Recognition (CVPR)}, pp.\  2523--2534, 2025.

\bibitem[Maher(2010)]{maher2010evaluating}
Mary~Lou Maher.
\newblock Evaluating creativity in humans, computers, and collectively intelligent systems.
\newblock In \emph{Proceedings of the 1st DESIRE Network Conference on Creativity and Innovation in Design}, pp.\  22--28, 2010.

\bibitem[Mou et~al.(2025)Mou, Wu, Wu, Guo, Zhang, Cheng, Luo, Ding, Zhang, Li, et~al.]{mou2025dreamo}
Chong Mou, Yanze Wu, Wenxu Wu, Zinan Guo, Pengze Zhang, Yufeng Cheng, Yiming Luo, Fei Ding, Shiwen Zhang, Xinghui Li, et~al.
\newblock Dreamo: A unified framework for image customization.
\newblock In \emph{Proceedings of the SIGGRAPH Asia 2025 Conference Papers}, 2025.

\bibitem[{OpenAI}(2025)]{openai2023chatgpt}
{OpenAI}.
\newblock Chatgpt: Optimizing language models for dialogue.
\newblock 2025.
\newblock URL \url{https://www.openai.com}.
\newblock Accessed: 2025-05-07.

\bibitem[Oquab et~al.(2024)Oquab, Darcet, Moutakanni, Vo, Szafraniec, Khalidov, Fernandez, Haziza, Massa, El-Nouby, Assran, Ballas, Galuba, Howes, Huang, Li, Misra, Rabbat, Sharma, Synnaeve, Xu, Jegou, Mairal, Labatut, Joulin, and Bojanowski]{oquab2024dinov2}
Maxime Oquab, Timothée Darcet, Théo Moutakanni, Huy Vo, Marc Szafraniec, Vasil Khalidov, Pierre Fernandez, Daniel Haziza, Francisco Massa, Alaaeldin El-Nouby, Mahmoud Assran, Nicolas Ballas, Wojciech Galuba, Russell Howes, Po-Yao Huang, Shang-Wen Li, Ishan Misra, Michael Rabbat, Vasu Sharma, Gabriel Synnaeve, Hu~Xu, Hervé Jegou, Julien Mairal, Patrick Labatut, Armand Joulin, and Piotr Bojanowski.
\newblock Dinov2: Learning robust visual features without supervision.
\newblock \emph{Transactions on Machine Learning Research (TMLR)}, 2024.

\bibitem[Radford et~al.(2021)Radford, Kim, Hallacy, Ramesh, Goh, Agarwal, Sastry, Askell, Mishkin, Clark, et~al.]{radford2021learning}
Alec Radford, Jong~Wook Kim, Chris Hallacy, Aditya Ramesh, Gabriel Goh, Sandhini Agarwal, Girish Sastry, Amanda Askell, Pamela Mishkin, Jack Clark, et~al.
\newblock Learning transferable visual models from natural language supervision.
\newblock In \emph{Proceedings of the International Conference on Machine Learning (ICML)}, pp.\  8748--8763, 2021.

\bibitem[Ren et~al.(2024)Ren, Liu, Zeng, Lin, Li, Cao, Chen, Huang, Chen, Yan, et~al.]{ren2024grounded}
Tianhe Ren, Shilong Liu, Ailing Zeng, Jing Lin, Kunchang Li, He~Cao, Jiayu Chen, Xinyu Huang, Yukang Chen, Feng Yan, et~al.
\newblock Grounded sam: Assembling open-world models for diverse visual tasks.
\newblock \emph{arXiv preprint arXiv:2401.14159}, 2024.

\bibitem[Richardson et~al.(2024)Richardson, Goldberg, Alaluf, and Cohen-Or]{Richardson2024conceptlab}
Elad Richardson, Kfir Goldberg, Yuval Alaluf, and Daniel Cohen-Or.
\newblock Conceptlab: Creative concept generation using vlm-guided diffusion prior constraints.
\newblock \emph{ACM Transactions on Graphics (TOG)}, 43\penalty0 (3):\penalty0 1--14, 2024.

\bibitem[Roberts et~al.(2022)Roberts, Chung, Levskaya, Mishra, Bradbury, Andor, Narang, Lester, Gaffney, Mohiuddin, Hawthorne, Lewkowycz, Salcianu, van Zee, Austin, Goodman, Soares, Hu, Tsvyashchenko, Chowdhery, Bastings, Bulian, Garcia, Ni, Chen, Kenealy, Clark, Lee, Garrette, Lee-Thorp, Raffel, Shazeer, Ritter, Bosma, Passos, Maitin-Shepard, Fiedel, Omernick, Saeta, Sepassi, Spiridonov, Newlan, and Gesmundo]{roberts2022t5x}
Adam Roberts, Hyung~Won Chung, Anselm Levskaya, Gaurav Mishra, James Bradbury, Daniel Andor, Sharan Narang, Brian Lester, Colin Gaffney, Afroz Mohiuddin, Curtis Hawthorne, Aitor Lewkowycz, Alex Salcianu, Marc van Zee, Jacob Austin, Sebastian Goodman, Livio~Baldini Soares, Haitang Hu, Sasha Tsvyashchenko, Aakanksha Chowdhery, Jasmijn Bastings, Jannis Bulian, Xavier Garcia, Jianmo Ni, Andrew Chen, Kathleen Kenealy, Jonathan~H. Clark, Stephan Lee, Dan Garrette, James Lee-Thorp, Colin Raffel, Noam Shazeer, Marvin Ritter, Maarten Bosma, Alexandre Passos, Jeremy Maitin-Shepard, Noah Fiedel, Mark Omernick, Brennan Saeta, Ryan Sepassi, Alexander Spiridonov, Joshua Newlan, and Andrea Gesmundo.
\newblock Scaling up models and data with $\texttt{t5x}$ and $\texttt{seqio}$.
\newblock \emph{arXiv preprint arXiv:2203.17189}, 2022.

\bibitem[Sheynin et~al.(2024)Sheynin, Polyak, Singer, Kirstain, Zohar, Ashual, Parikh, and Taigman]{sheynin2024emu}
Shelly Sheynin, Adam Polyak, Uriel Singer, Yuval Kirstain, Amit Zohar, Oron Ashual, Devi Parikh, and Yaniv Taigman.
\newblock Emu edit: Precise image editing via recognition and generation tasks.
\newblock In \emph{Proceedings of the IEEE/CVF Conference on Computer Vision and Pattern Recognition (CVPR)}, pp.\  8871--8879, 2024.

\bibitem[Shoemake(1985)]{shoemake1985animating}
Ken Shoemake.
\newblock Animating rotation with quaternion curves.
\newblock In \emph{Proceedings of the 12th annual conference on Computer graphics and interactive techniques}, pp.\  245--254, 1985.

\bibitem[Tang et~al.(2023)Tang, Yu, and Song]{Tang2023master}
H.~Tang, L.~Yu, and J.~Song.
\newblock Master: Meta style transformer for controllable zero-shot and few-shot artistic style transfer.
\newblock In \emph{Proceedings of the IEEE/CVF Conference on Computer Vision and Pattern Recognition (CVPR)}, 2023.

\bibitem[Teukolsky et~al.(1992)Teukolsky, Flannery, Press, and Vetterling]{teukolsky1992numerical}
Saul~A Teukolsky, Brian~P Flannery, W~Press, and W~Vetterling.
\newblock Numerical recipes in c.
\newblock \emph{SMR}, 693\penalty0 (1):\penalty0 59--70, 1992.

\bibitem[Wang et~al.(2024)Wang, Ostashev, Fang, Tulyakov, and Aberman]{MoA2024}
Kuan-Chieh Wang, Daniil Ostashev, Yuwei Fang, Sergey Tulyakov, and Kfir Aberman.
\newblock Moa: Mixture-of-attention for subject-context disentanglement in personalized image generation.
\newblock In \emph{Proceedings of the SIGGRAPH Asia 2024 Conference Papers}, 2024.

\bibitem[Wang et~al.(2023)Wang, Que, Chen, Li, Li, and Yang]{wang2023creative}
Renke Wang, Guimin Que, Shuo Chen, Xiang Li, Jun Li, and Jian Yang.
\newblock Creative birds: Self-supervised single-view 3d style transfer.
\newblock In \emph{Proceedings of the IEEE/CVF International Conference on Computer Vision (ICCV)}, pp.\  8775--8784, 2023.

\bibitem[Xiao et~al.(2025)Xiao, Wang, Zhou, Yuan, Xing, Yan, Li, Wang, Huang, and Liu]{xiao2024omnigen}
Shitao Xiao, Yueze Wang, Junjie Zhou, Huaying Yuan, Xingrun Xing, Ruiran Yan, Chaofan Li, Shuting Wang, Tiejun Huang, and Zheng Liu.
\newblock Omnigen: Unified image generation.
\newblock In \emph{Proceedings of the IEEE/CVF conference on Computer Vision and Pattern Recognition (CVPR)}, pp.\  13294--13304, 2025.

\bibitem[Xiong et~al.(2025{\natexlab{a}})Xiong, Wang, Guo, Ye, Fan, Gu, Huang, and Li]{xiong2025llava}
Tianyi Xiong, Xiyao Wang, Dong Guo, Qinghao Ye, Haoqi Fan, Quanquan Gu, Heng Huang, and Chunyuan Li.
\newblock Llava-critic: Learning to evaluate multimodal models.
\newblock In \emph{Proceedings of the Computer Vision and Pattern Recognition Conference (CVPR)}, pp.\  13618--13628, 2025{\natexlab{a}}.

\bibitem[Xiong et~al.(2024)Xiong, dong Zhang, Chen, Chen, Li, Sun, Yang, and Li]{xiong2024novel}
Zeren Xiong, Ze~dong Zhang, Zikun Chen, Shuo Chen, Xiang Li, Gan Sun, Jian Yang, and Jun Li.
\newblock Novel object synthesis via adaptive text-image harmony.
\newblock In \emph{Proceedings of the Advances in Neural Information Processing Systems (NeurIPS)}, pp.\  139085--139113, 2024.

\bibitem[Xiong et~al.(2025{\natexlab{b}})Xiong, Chen, Zhang, Li, Tai, Yang, and Li]{xiong2025category}
Zeren Xiong, Zikun Chen, Zedong Zhang, Xiang Li, Ying Tai, Jian Yang, and Jun Li.
\newblock Category-aware 3d object composition with disentangled texture and shape multi-view diffusion.
\newblock In \emph{Proceedings of the ACM International Conference on Multimedia (ACMMM)}, 2025{\natexlab{b}}.

\bibitem[Zhang et~al.(2024)Zhang, Wei, Wu, Zhang, Zhang, Lei, and Li]{zhang2024compositional}
Xulu Zhang, Xiao-Yong Wei, Jinlin Wu, Tianyi Zhang, Zhaoxiang Zhang, Zhen Lei, and Qing Li.
\newblock Compositional inversion for stable diffusion models.
\newblock In \emph{Proceedings of the AAAI Conference on Artificial Intelligence (AAAI)}, number~7, pp.\  7350--7358, 2024.

\bibitem[Zhang et~al.(2023)Zhang, Wang, and Li]{Zhang2023inversion}
Y.~Zhang, L.~Wang, and H.~Li.
\newblock Inversion-based style transfer with diffusion models.
\newblock In \emph{Proceedings of the IEEE/CVF Conference on Computer Vision and Pattern Recognition (CVPR)}, pp.\  10146--10156, 2023.

\bibitem[Zhao et~al.(2024)Zhao, Bai, Zhang, Zhang, Zhang, Xu, Chen, Timofte, and Van~Gool]{zhao2024equivariant}
Zixiang Zhao, Haowen Bai, Jiangshe Zhang, Yulun Zhang, Kai Zhang, Shuang Xu, Dongdong Chen, Radu Timofte, and Luc Van~Gool.
\newblock Equivariant multi-modality image fusion.
\newblock In \emph{Proceedings of the IEEE/CVF Conference on Computer Vision and Pattern Recognition (CVPR)}, pp.\  25912--25921, 2024.

\bibitem[Zheng et~al.(2024)Zheng, Zhou, Huang, Hou, Li, Xu, and Zhao]{zheng2024probing}
Naishan Zheng, Man Zhou, Jie Huang, Junming Hou, Haoying Li, Yuan Xu, and Feng Zhao.
\newblock Probing synergistic high-order interaction in infrared and visible image fusion.
\newblock In \emph{Proceedings of the IEEE/CVF Conference on Computer Vision and Pattern Recognition (CVPR)}, pp.\  26384--26395, 2024.

\bibitem[Zhou et~al.(2025)Zhou, Shen, and Wang]{zhou2025freeblend}
Yufan Zhou, Haoyu Shen, and Huan Wang.
\newblock Freeblend: Advancing concept blending with staged feedback-driven interpolation diffusion.
\newblock \emph{arXiv preprint arXiv:2502.05606}, 2025.

\bibitem[Zou et~al.(2025)Zou, Shen, Bouganis, and Zhao]{zou2025mcig}
Xiandong Zou, Mingzhu Shen, Christos-Savvas Bouganis, and Yiren Zhao.
\newblock Cached multi-lora composition for multi-concept image generation.
\newblock In \emph{Proceedings of the International Conference on Learning Representations (ICLR)}, 2025.

\end{thebibliography}
\bibliographystyle{iclr2026_conference}

\clearpage
\newpage
\appendix

\section*{Supplementary Materials} % Or \appendix followed by \section{Supplementary Materials} if using acmart.cls

This supplementary material provides additional technical details and extended results to support the main paper. We begin in \textbf{Section~\ref{sec:Discuss}} with two key discussions: the necessity of adjusting \(\beta_1\) and \(\beta_2\) in our hierarchical parameter search, and a quantitative comparison of BNoise fusion strategies—concatenation versus interpolation.
\textbf{Section~\ref{sec:sup_Dataset}} describes the construction of our proposed IIOF benchmark dataset, including the criteria for category selection and object pairing strategies. \textbf{Section~\ref{sec:sup_user_stu}} presents a comprehensive user study, providing human preference validation of our fusion results. In \textbf{Section~\ref{sec:limitation}}, we outline the current limitations of our method, discuss remaining challenges, and suggest possible directions for future improvement. 
\textbf{Section~\ref{sec:Algorithm}} details the full inference pipeline of our VMDiff framework. Finally, \textbf{Section~\ref{sec:more_results}} showcases extensive qualitative results, further demonstrating the effectiveness and generalization ability of our method across diverse fusion scenarios.

% \textbf{Section~\ref{sec:State_LLM}} contains our formal statement on the use of LLMs in this work, in accordance with ICLR policy.

\section{Additional discussions.}
\label{sec:Discuss}

\begin{figure}[h]
    \centering
    \includegraphics[width=0.85\linewidth]{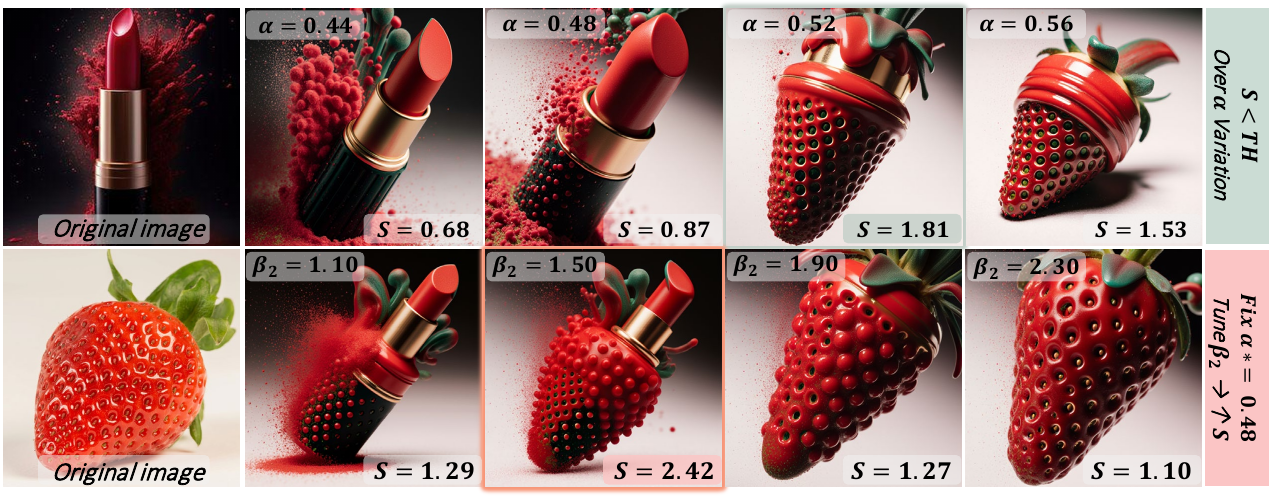}
    \caption{Illustration of our hierarchical parameter adjustment. The top row shows results from searching $\alpha$; the bottom row refines the fusion by fixing $\alpha$ and adjusting $\beta_2$. \textbf{Consistent with Sec.~\ref{sec:EAA}, once the overall score $S$ exceeds the acceptance threshold $T_h\!=\!2.4$, the fusion becomes visually coherent and balanced}; when $\alpha$-only optimization underperforms, the second-stage $\beta_2$ refinement raises $S$ above the threshold.}
    \label{fig:wht_alpha_beta}
    \vskip -0.1in
\end{figure}
%%%%%%%%%%%%%%%%%%%%
\textit{\textbf{Discussion on the necessity of adjusting $\beta_1,\beta_2$.}} As shown in Fig.~\ref{fig:wht_alpha_beta}, global optimization over $\alpha$ alone occasionally fails to yield well-fused results. To mitigate this, we first fix \(\alpha^*\) (corresponding to the best similarity score in Eq. \eqref{eq:score}) and then perform a local refinement by optimizing \(\beta_1\), \(\beta_2\). This adjustment allows the model to precisely calibrate the noise contribution of each object, enhancing both visual coherence and semantic balance in the final output.

\textit{\textbf{Discussion on BNoise.}} As shown in Table~\ref{tab:noise_ablatuion} on the IIOF dataset, Ours (Concat before inversion) achieves state-of-the-art performance on most metrics. Although it ranks second on the LC metric, its substantial advantage on SS, demonstrates that concatenation more effectively preserves and integrates complementary information from both inputs. In summary, concatenation before inversion yields superior visual quality and semantic faithfulness by retaining fine-grained details and guiding a more coherent denoising pathway, compared with either form of interpolation.

%%%%%%%%%%%%%%%%%%%%%%%%%%%%%%%%%%%%%%%%%%%%%%%%%%%%%%%%%%%%%%%%%%%%%%%%%%%%%%
\begin{table}[h]
\vskip -0.2in
\centering
\setlength{\tabcolsep}{7pt}
\renewcommand{\arraystretch}{0.97}
\caption{Quantitative Evaluation of BNoise Fusion: Concatenation vs. Interpolation.}
\label{tab:noise_ablatuion}
%\vskip -0.1in
\resizebox{0.95\linewidth}{!}{
\begin{tabular}{c||c|c|c|c|c|c|c|c}
\toprule[1.2pt]
\addlinespace[1pt]  
Models & $\mathrm{VQA}^{\mathrm{SA}}_{\mathrm{T5}}$$\uparrow$      & $\mathrm{VQA}^{\mathrm{SCE}}_{\mathrm{T5}}$$\uparrow$     & $\mathrm{LC}^{\mathrm{SA}}$ $\uparrow$  &$\mathrm{LC}^{\mathrm{SCE}}$$\uparrow$   &  $\mathrm{VQA}^{\mathrm{SA}}_{\mathrm{LLaVA}}\uparrow$  & $\mathrm{VQA}^{\mathrm{SCE}}_{\mathrm{LLaVA}}$$\uparrow$ &$SS\uparrow$  & $B$sim$\downarrow$\\ \hline
 Random noise           &0.497  & 0.438  & 7.261 & 
7.077 &
0.287     & 0.314 & 1.570 & 0.682\\
Interp Before Inversion        & 0.504 & 0.441  & \textbf{7.439} & \textbf{7.390} & 0.293      & 0.321  & 1.551 & \textbf{0.678} \\
Interp After Inversion  & 0.486 &  0.430   & 7.278  & 7.112 & 0.283   & 0.311  &1.532  &0.712 \\
\textbf{Ours(Concat Before Inversion)} &\textbf{ 0.508} &  \textbf{0.442}   & 7.426  & 7.291 & \textbf{0.298} & \textbf{0.325}  &\textbf{1.586} &
0.693 \\
\bottomrule
\end{tabular}}
\vskip -0.1in
\end{table}
%%%%%%%%%%%%%%%%%%%%%%%%%%%%%%%%%%%%%%%%%%%%%%%%%%%%%%%%%%%%%%%%%%%%%%%%%%%%%% 

\section{Datasets}
\label{sec:sup_Dataset}
To systematically evaluate our fusion framework, we construct a comprehensive benchmark dataset named \textbf{IIOF} (Image-Image Object Fusion), specifically tailored for assessing diverse and semantically rich visual concept mixing.

We meticulously selected \textbf{40 distinct object categories}, strategically organized into four semantic groups: \textit{Animals}, \textit{Fruits}, \textit{Artificial Objects}, and \textit{Character Figurines}. Each group comprises 10 unique classes, a design choice that ensures both intra-group consistency and ample inter-group diversity. A complete list of all selected categories is provided in Table~\ref{tab:IIOF_object_categories}.

\begin{figure}[htbp]
    \centering
    \includegraphics[width=0.97\linewidth]{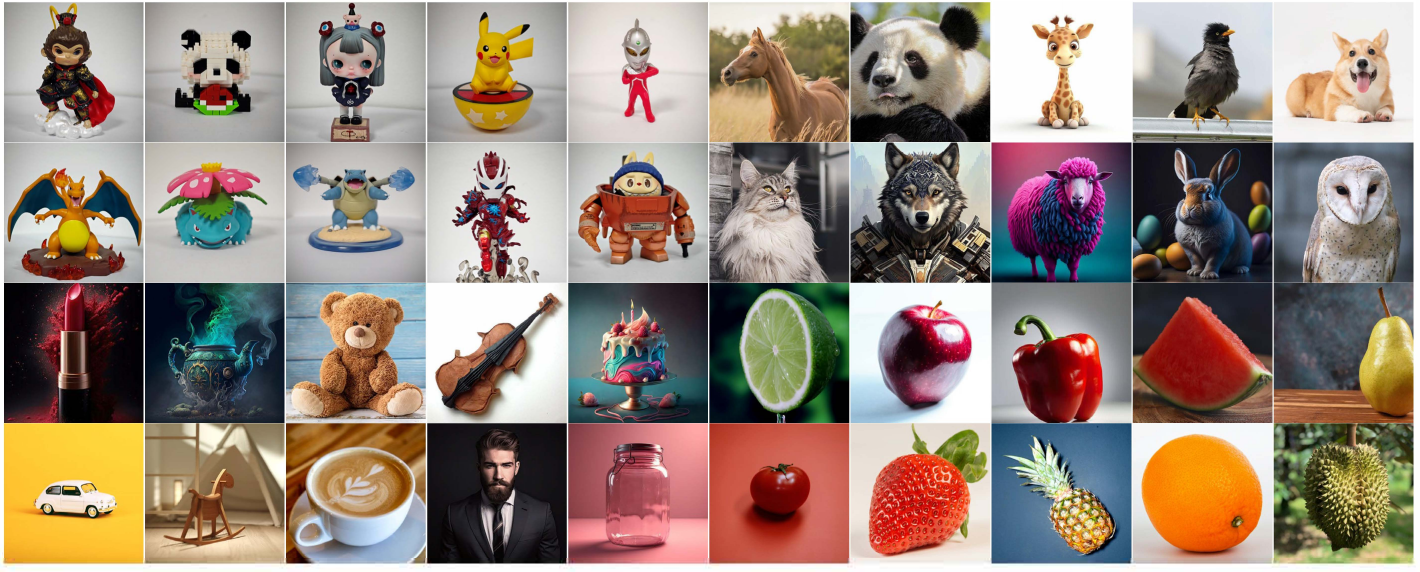}
    \vskip -0.1in
    \caption{Original Object Image Set.}
    \label{fig:dataset}
\end{figure}

For each chosen class, we sourced one high-quality, representative image. The majority of these images were obtained from established public benchmarks such as PIE-Bench~\cite{ju2023direct} and popular stock image platforms like Pexels\footnote{\url{https://www.pexels.com/}}. Recognizing the scarcity of high-quality, publicly available data for character figurines, we self-captured these images under controlled conditions, ensuring consistent lighting and resolution to maintain visual quality and diversity across the dataset. Figure~\ref{fig:dataset} showcases all the selected images, providing a visual overview of the dataset's content. Additionally, each selected image is paired with its corresponding \textbf{textual category name}, as detailed in Table~\ref{tab:IIOF_object_categories}, to facilitate evaluations for prompt-based fusion methods.
%%%%%%%%%%%%%%%%%%%%
\begin{wraptable}{r}{0.6\textwidth}
    \vspace{-5mm} % 可调整与上方段落距离
    \centering
    \setlength{\tabcolsep}{6pt}
    \renewcommand{\arraystretch}{1.05}
    \caption{List of Objects in the IIOF Dataset by Category.}
    \label{tab:IIOF_object_categories}
    \begin{tabular}{>{\raggedright}m{1.5cm}||>{\raggedright\arraybackslash}m{5.4cm}}
        \Xhline{1.2pt}
        \textbf{Category} & \textbf{Object Names} \\
        \hline
        Animals & wolf, panda, owl, rabbit, horse, giraffe, corgi, cat, bird, sheep \\
        \hline
        Fruits & apple, orange, strawberry, durian, lime, pear, pineapple, watermelon, tomato, pepper \\
        \hline
        Artificial Objects & lipstick, violin, coffee cup, rocking horse, glass jar, car, teapot, cake, man, teddy bear \\
        \hline
        Character Figurines & iron man figurine, monkey king figurine, doll figurine, pikachu figurine, charizard figurine, ultraman figurine, astronaut figurine, venusaur figurine, panda figurine, squirtle figurine \\
        \Xhline{1.2pt}
    \end{tabular}
    \vspace{-3mm} % 可调整与下方段落距离
\end{wraptable}
%%%%%%%%%%%%%%%%%%%%%%%%%%%%
Initially, we derived \textbf{780 unique image pairs} by combining each of the 40 objects with every other object once, without considering input order. However, to ensure a comprehensive evaluation and enable fair comparison across all methods, particularly those sensitive to input order (e.g., ATIH~\cite{xiong2024novel}), we further expanded IIOF to include \textbf{all possible ordered pairs} among the 40 categories. This expansion yielded a total of \textbf{1,560 image pairs}, where each combination $(A, B)$ is present alongside its reverse $(B, A)$. This exhaustive pairing strategy allows us to rigorously assess fusion performance across a wide spectrum of semantic relationships---ranging from semantically close concepts to challenging distant combinations, such as fusing a 'violin' with a 'panda' or a 'horse' with 'lipstick'. This also critically highlights our model's ability to generalize and compose novel concepts effectively across diverse domains.

%%%%%%%%%%%%%%%%%%%%%%%%%%%
\section{User Study}
\label{sec:sup_user_stu}

\begin{wrapfigure}{r}{0.77\textwidth}
\vskip -0in
   \centering
   \includegraphics[width=0.97\linewidth]{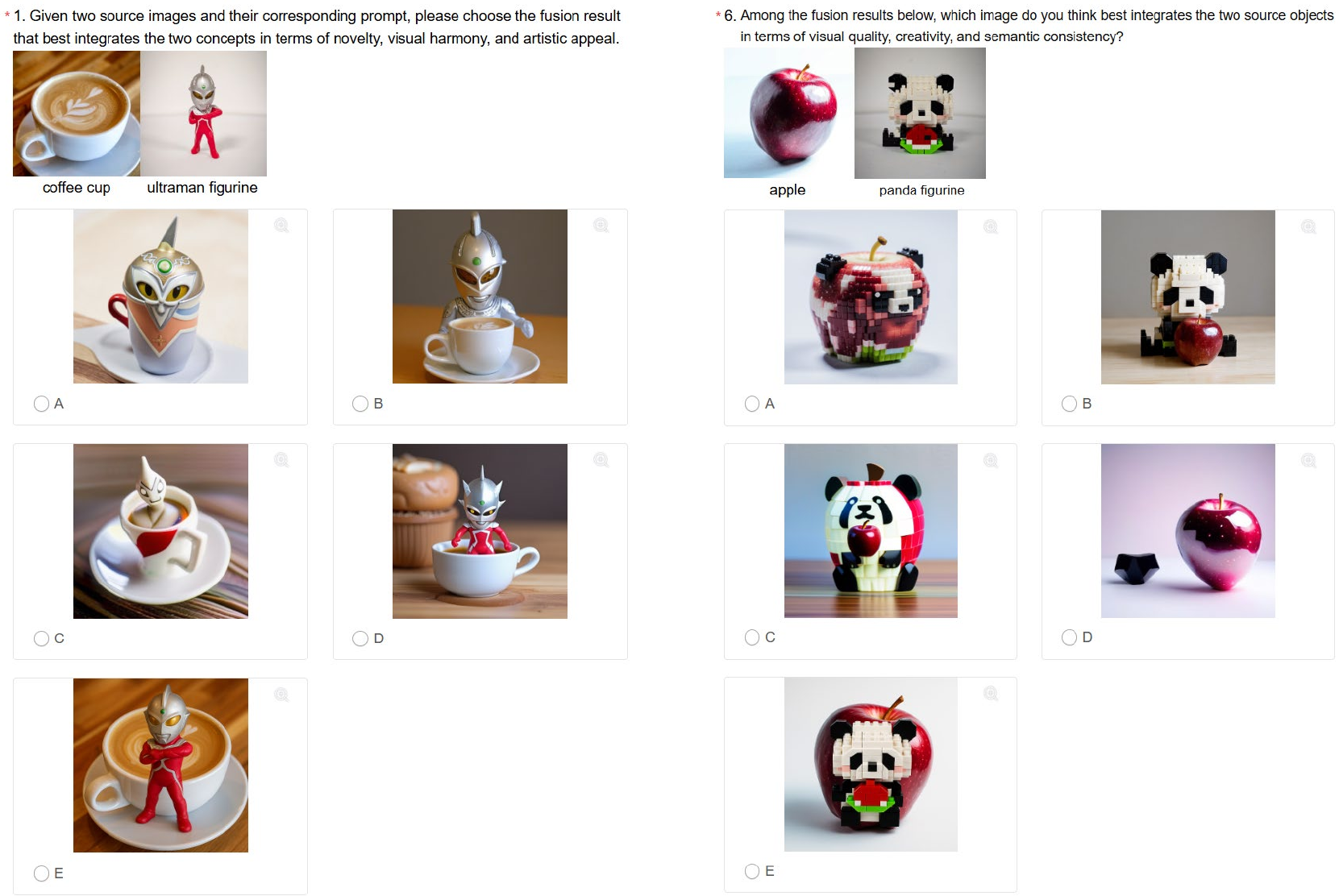}
   \vskip -0.1in
   \caption{An example of a user study comparing various multi-concept generation, mixing and image editing methods.}
   \label{fig:User_study_problem_1}
   \vskip -0.05in
\end{wrapfigure}

To evaluate the perceptual quality and human preference for the novel images generated by our fusion framework, we conducted two user studies. These studies assessed our method, \textbf{VMDiff}, against state-of-the-art baselines in two main categories: \textit{Multi-Concept Generation} methods and \textit{Mixing and Image Editing} methods. The overall vote distributions are visualized in Fig.~\ref{fig:user_study}, while detailed per-example preferences are presented in Table~\ref{tab:sup_votes1} and Table~\ref{tab:sup_votes2}. An example user study question for the \textit{Multi-Concept Generation} group and the \textit{Mixing and Image Editing} group are provided in Fig.~\ref{fig:User_study_problem_1}.
A total of 76 participants completed the survey, each evaluating 12 fused results (6 from each group), contributing a total of 912 votes. Participants were asked to select the fusion result that best integrated the given concepts in terms of visual quality, creativity, and semantic consistency.
As shown in Fig.~\ref{fig:user_study}, our method consistently received the highest number of votes in both evaluation groups. In the \textit{Mixing and Image Editing} category (left pie chart), VMDiff garnered a significant \textbf{397 votes (87.1\%)} of the total. This considerably surpassed other methods such as Stable Flow~\cite{avrahami2024stable} (5 votes, 1.1\%), ATIH~\cite{xiong2024novel} (34 votes, 7.5\%),  Conceptlab~\cite{Richardson2024conceptlab} (4 votes, 0.9\%) and FreeBlend~\cite{zhou2025freeblend} (16 votes, 3.5\%). For instance, as illustrated in Fig.~\ref{fig:User_study_problem_1}, for the ``astronaut figurine-monkey king figurine '' fusion, our method obtained 81.58\% of the votes, demonstrating its strong capability in seamlessly integrating distinct visual elements.
%%%%%%%%%%%%%%%%%%%
%%%%%%%%%%%%%%%%%%%
In the \textit{Multi-Concept Generation} category (right pie chart), \textbf{VMDiff} led with \textbf{307 votes (67.3\%)}, significantly outperforming GPT-4o~\cite{openai2023chatgpt}, which ranked second with 59 votes (12.9\%). Other baselines—DreamO (56 votes, 12.3\%), MIP-Adapter (17 votes, 3.7\%), and OmniGen (17 votes, 3.7\%)—received notably fewer votes. In the ``doll figurine–corgi'' case , VMDiff earned \textbf{78.95\%} of preferences. Even in more challenging cases like ``apple–panda figurine ''(see Fig.~\ref{fig:User_study_problem_1}), it maintained an edge with \textbf{75.00\%} over GPT-4o's \textbf{5.26\%}.
These results indicate that \textbf{VMDiff} better aligns with human preferences for visual coherence, creativity, and concept integration, consistently outperforming existing methods across diverse fusion scenarios.

\begin{table}[htbp]
\vskip -0.25in
\centering
\setlength{\tabcolsep}{7pt}
\renewcommand{\arraystretch}{1.1}
%\caption{User study with Image-to-3D methods.}
\caption{User study with multi-concept generation methods.}
%\vskip -0.1in
\resizebox{0.95\linewidth}{!}{
\begin{tabular}{c||c|c|c|c|c}
\toprule[1.2pt]
\addlinespace[1pt]    
\diagbox{image-image} & A(Our VMDiff) & B(DreamO) & C(MIP-Adapter) & D( OmniGen) & E(GPT-4o)\\ 
\hline
coffee cup-ultraman figurine  & 43(56.58\%) & 11(14.47\%) & 7(9.21\%) & 3(3.95\%) & 12(15.79\%) \\
sheep-car  & 57(75.00\%) & 4(5.26\%) & 1(1.32\%) & 2(2.63\%) & 12(15.79\%) \\
doll figurine-corgi  & 60(78.95\%) & 1(1.32\%) & 3(3.95\%) & 3(3.95\%) & 9(11.84\%) \\
lime-glass jar  & 45(59.21\%) & 22(28.95\%) & 1(1.32\%) & 0(0.00\%) & 8(10.53\%) \\
cake-owl  & 45(59.21\%) & 5(6.58\%) & 3(3.95\%) & 9(11.84\%) & 14(18.42\%) \\
apple-panda figurine  & 57(75.00\%) & 13(17.11\%) & 2(2.63\%) & 0(0.00\%) & 4(5.26\%) \\
\toprule[1.2pt]
\addlinespace[1pt]   

\end{tabular}}

\label{tab:sup_votes1}
\end{table}

\begin{table}[htbp]
\vskip -0.32in
\centering
\setlength{\tabcolsep}{7pt}
\renewcommand{\arraystretch}{1.1}
\caption{User study with mixing and image editing methods.}
%\vskip -0.1in
\resizebox{0.95\linewidth}{!}{
\begin{tabular}{c||c|c|c|c|c}
\toprule[1.2pt]
\addlinespace[1pt]   
\diagbox{image-image} & A(Our VMDiff)& B(Stable Flow)  & C(ATIH) & D(Conceptlab) & E(FreeBlend) \\ 
\hline
astronaut figurine-monkey king figurine  & 62(81.58\%) & 2(2.63\%) & 7(9.21\%) & 1(1.32\%) & 4(5.26\%)\\
man-pikachu figurine  & 68(89.47\%) & 0(0.00\%) & 4(5.26\%) & 1(1.32\%) & 3(3.95\%)\\
doll figurine-panda & 62(81.58\%) & 0(0.00\%) & 13(17.11\%) & 1(1.32\%) 
& 0(0.00\%)\\
iron man figurine-charizard figurine  & 69(90.79\%) & 3(3.95\%) & 3(3.95\%) & 0(0.00\%) & 1(1.32\%)\\
squirtle-wolf  & 66(86.84\%) & 0(0.00\%) & 4(5.26\%) & 1(1.32\%) & 5(6.58\%) \\
ultraman figurine-venusaur figurine  & 70(92.11\%) & 0(0.00\%) & 3(3.95\%) & 0(0.00\%) & 3(3.95\%) \\
\toprule[1.2pt]
\addlinespace[1pt]   
\end{tabular}}
\label{tab:sup_votes2}
\end{table}

%%%%%%%%%%%%%%%%%%
%%%%%%%%%%%%%%%

\section{Limitation}
\label{sec:limitation}

Our method effectively fuses two input images into a coherent hybrid object that captures broad conceptual information; however, it has two main limitations. First, inference relies on iterative optimization, which increases computational cost and latency (Table~\ref{tab:eaa_runtime}).
%%%%%%%%%%%%%%%%%%%
\begin{wrapfigure}{r}{0.65\textwidth}
\vskip -0.1in
   \centering
   \includegraphics[width=0.95\linewidth]{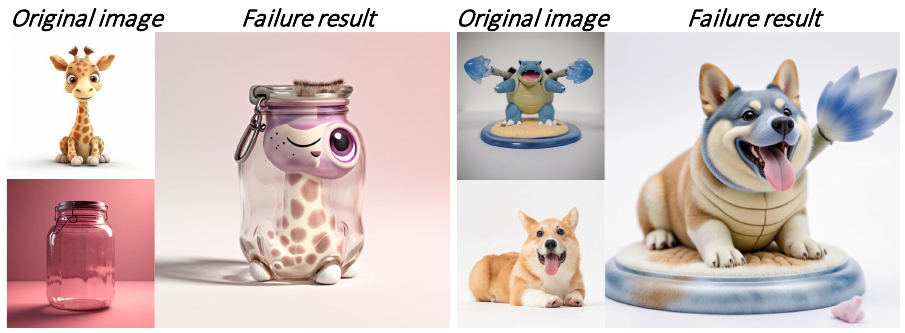}
   \vskip -0.1in
   \caption{Examples of failure cases where our method produces fused outputs with suboptimal semantic or stylistic coherence.}
   \label{fig:Fail_case}
   \vskip -0.05in
\end{wrapfigure}
%%%%%%%%%%%%%%%%%%%
A promising remedy is to train a lightweight prediction/refinement module that guides the fusion in a single forward pass, thereby reducing runtime while maintaining—or even improving—visual quality and semantic balance. Second, in a small fraction of cases the fused outputs do not fully align with human preferences (Fig.~\ref{fig:Fail_case}), exhibiting semantic inconsistencies or stylistic imbalance. Although repeated noise resampling and selection can mitigate these failures, this heuristic is  limited controllability. In future work, we will pursue more controllable, preference-aligned fusion via explicit human feedback, aesthetic priors, or learned alignment objectives, enabling results that more reliably reflect human intent and aesthetics.

%%%%%%%%%%%%%%%%%%%%%%%%%%%%
\begin{table}[htbp]
\vskip -0.2in
\centering
\caption{Runtime comparison across methods.}
\label{tab:eaa_runtime}
\begin{tabular}{lcc}
\toprule
\textbf{Methods} & \textbf{Avg. Time / Pair}  \\
\midrule
Ours  & 2 min 46 sec \\
ATIH~ & 10 sec  \\
Stable Flow & 27 sec \\
Conceptlab & 13 min 45 sec \\
FreeCustom & 22 sec  \\
OmniGen & 53 sec  \\
Freeblend & 12 sec   \\
MIP-Adapter & 12 sec \\
DreamO & 8 sec \\
\bottomrule
\end{tabular}
\end{table}

\begin{algorithm}[htbp]
\caption{VMDiff with Efficient Adaptive Adjustment (VMDiff-EAA)}
\label{alg:vmdiff_main}
\KwIn{images $I_1,I_2$, labels $T_1,T_2$, prompt $P_G$, threshold $TH$, max rounds $K$}
\KwOut{fused image $I^*$ and parameters $\theta^*=\{\alpha^*,\beta_1^*,\beta_2^*,\epsilon^*\}$}

% ----- Initialization -----
Compute embeddings $z_1=\mathcal{E}_I(I_1),\ z_2=\mathcal{E}_I(I_2),\ z_p=\mathcal{E}_T(P_G)$\;
Initialize $\alpha=0.5,\ \beta_1=\beta_2=1.0$;\quad $S_{\text{best}}=-\infty,\ \theta_{\text{best}}=\varnothing$\;

% ----- Main Loop -----
\For{$k=1$ \KwTo $K$}{
  Sample noise $\epsilon\sim\mathcal{N}(0,I)$\;

  % ----- RefineNoise -----
  $z_{\text{SCat}}=\text{concat}(\beta_1z_1,\beta_2z_2)$,\ $x_T=\epsilon$\;
  \For{$t=T$ \KwTo $t_{\text{den}}$}{
    $x_{t-1}=x_t-(\sigma_t-\sigma_{t-1})v_\phi(x_t,t,z_{\text{SCat}},\gamma_{\text{den}},z_p)$
  }
  \For{$t=t_{\text{den}}$ \KwTo $T$}{
    $x_{t+1}=\hat{x}_t+(\sigma_{t+1}-\sigma_t)v_\phi(\hat{x}_t,t,z_{\text{SCat}},\gamma_{\text{inv}},z_p)$
  }
  $\epsilon_r=\hat{x}_T$\;

  % ----- GoldenSearch for alpha -----
  $\alpha^*=\text{GoldenSearch}(\alpha\in[0,1], f(\alpha)=S(\alpha,\beta_1,\beta_2,\epsilon_r))$\;

  $(S,S_{I_1},S_{I_2},S_{T_1},S_{T_2})=\text{Score}(\alpha^*,\beta_1,\beta_2,\epsilon_r)$\;
  \If{$S>S_{\text{best}}$}{ $S_{\text{best}}=S$;\ $\theta_{\text{best}}=\{\alpha^*,\beta_1,\beta_2,\epsilon_r\}$ }
  \If{$S \ge TH$}{ \Return $I(\theta^*),\ \theta^*$ }

  % ----- Adaptive update of betas -----
  $S_1=S_{I_1}+S_{T_1}$,\ $S_2=S_{I_2}+S_{T_2}$\;
  \eIf{$S_1>S_2$}{
    $\beta_2^*=\text{GoldenSearch}(\beta_2\in[\beta_{\min},\beta_{\max}], f(\beta_2))$
  }{
    $\beta_1^*=\text{GoldenSearch}(\beta_1\in[\beta_{\min},\beta_{\max}], f(\beta_1))$
  }

  $(S',\cdot)=\text{Score}(\alpha^*,\beta_1^*,\beta_2^*,\epsilon_r)$\;
  \If{$S'>S_{\text{best}}$}{ $S_{\text{best}}=S'$;\ $\theta_{\text{best}}=\{\alpha^*,\beta_1^*,\beta_2^*,\epsilon_r\}$ }
  \If{$S' \ge TH$}{
    % ----- MixingDenoise -----
    Normalize $z_1,z_2$ and compute spherical interpolation $z_{\text{SInp}}(\alpha^*)$\;
    $x_T=\epsilon_r$\;
    \For{$t=T$ \KwTo $0$}{
      $x_{t-1}=x_t-(\sigma_t-\sigma_{t-1})v_\phi(x_t,t,z_{\text{SInp}}(\alpha^*),\gamma_{\text{gen}},z_p)$
    }
    $I=\mathcal{D}(x_0)$;\quad \Return $I,\ \theta^*$
  }
}

% ----- Fallback -----
\If{$\theta_{\text{best}}\neq\varnothing$}{
  Decode best parameters $\theta_{\text{best}}$ via MixingDenoise\;
  \Return $I,\ \theta_{\text{best}}$
}
\Return $\varnothing$\;
\end{algorithm}
%%%%%%%%%%%%%%%

%%%%%%%%%%%%
\section{Algorithm}
%%%%%%%fake code%%%%%%%%

\label{sec:Algorithm}

Algorithm~\ref{alg:vmdiff_main} outlines the complete inference process of our proposed framework, \textbf{VMDiff}, which integrates a noise refinement step and an efficient adaptive adjustment (EAA) loop. Given two input images \( I_1, I_2 \) and their category labels \( T_1, T_2 \), we construct a prompt \( P_G \) and initialize the fusion parameters \( \theta = \{\alpha, \beta_1, \beta_2, \epsilon\} \).

The algorithm begins by sampling initial Gaussian noise \( \epsilon \), which is refined through a denoising-inversion procedure to produce a structure-aware latent representation \( \epsilon_r \). The core loop involves:

\begin{itemize}
    \item \textbf{Searching} for the optimal interpolation factor \( \alpha \) using Golden Section Search  to maximize the similarity score \( S(\theta) \).
    \item \textbf{Conditionally adjusting} the noise scaling factors \( \beta_1, \beta_2 \) when the current fusion score is below a threshold \( TH \), guiding the fusion toward balance between the two source objects.
    \item \textbf{Returning} a fused image \( I(\theta^*) \) once a satisfactory similarity score is achieved.
\end{itemize}

This design ensures a lightweight and interpretable optimization routine over a low-dimensional parameter space. The algorithm reliably produces perceptually and semantically coherent hybrid images, as validated in our experiments.

\section{More Results}
\label{sec:more_results}

\begin{figure}[h]
    \centering
    \includegraphics[width=0.87\linewidth]{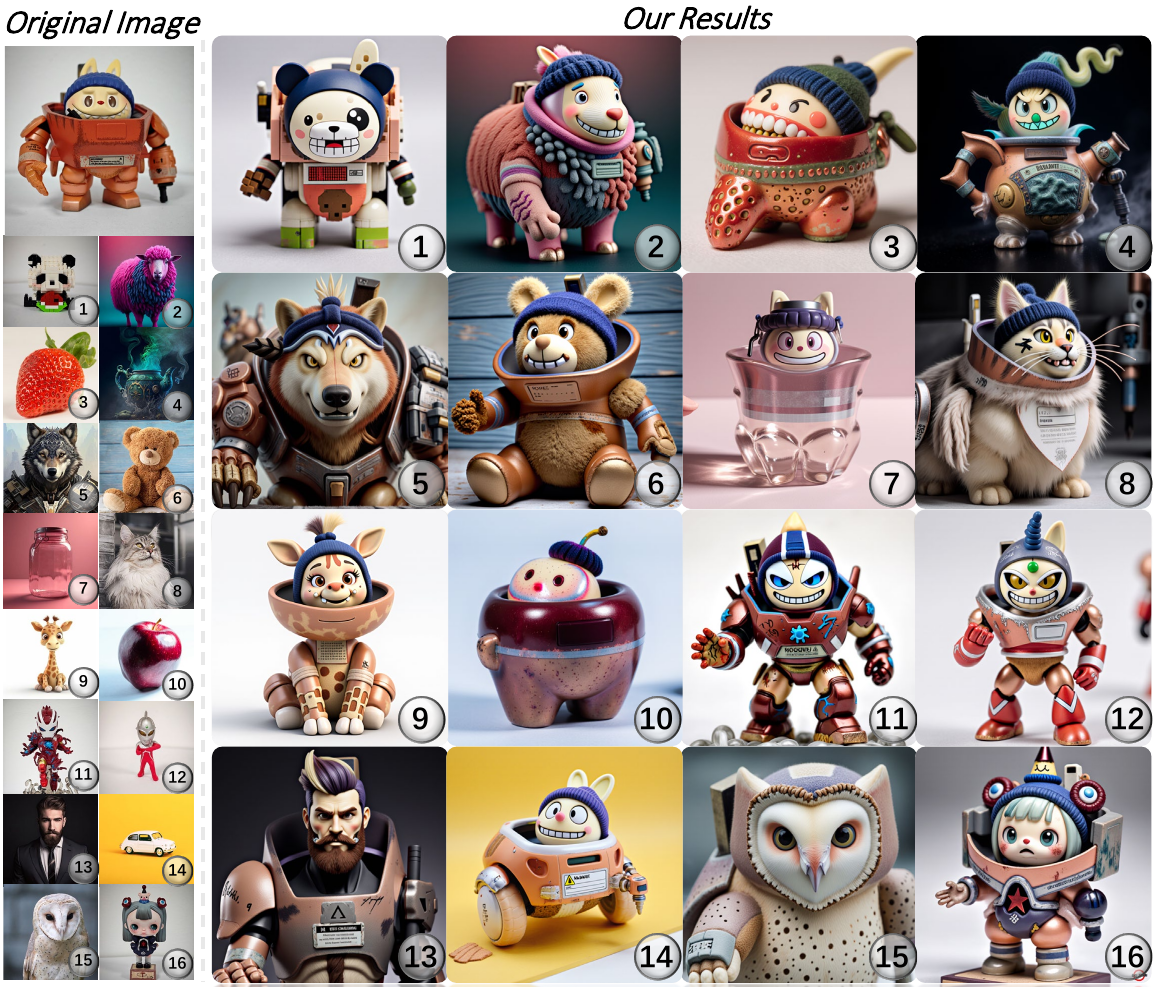}
    \vskip -0.1in
    \caption{\textbf{More Results.} The primary source (\textit{astronaut  figurine}, top-left) is fused with secondary inputs (left column), with results shown on the right.}
    \label{fig:More_result_1}
\end{figure}

\begin{figure}[h]
    \centering
    \includegraphics[width=0.87\linewidth]{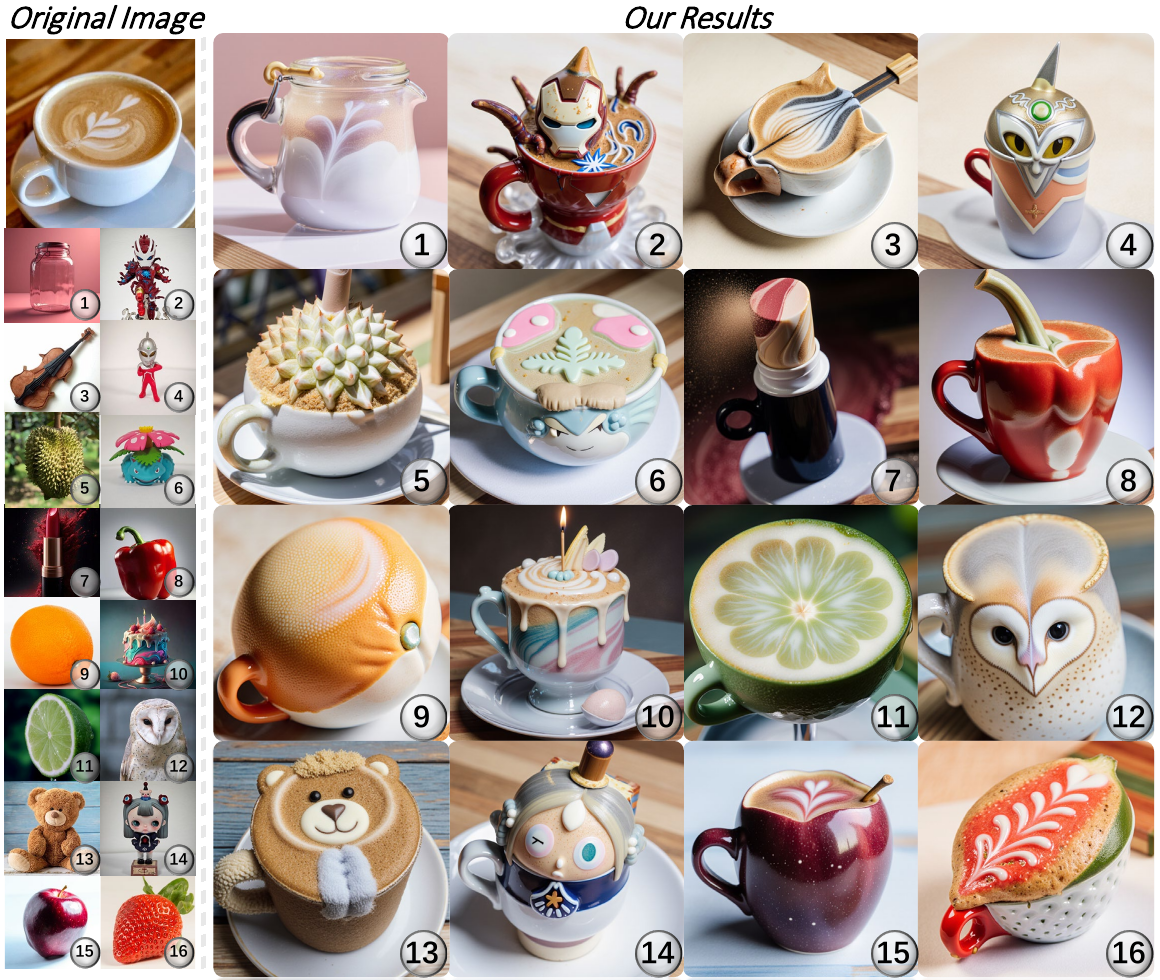}
    \vskip -0.1in
    \caption{\textbf{More Results.} The primary source (\textit{coffee cup}, top-left) is fused with secondary inputs (left column), with results shown on the right.}
    \label{fig:More_result_2}
\end{figure}

\begin{figure}[h]
    \centering
    \includegraphics[width=0.87\linewidth]{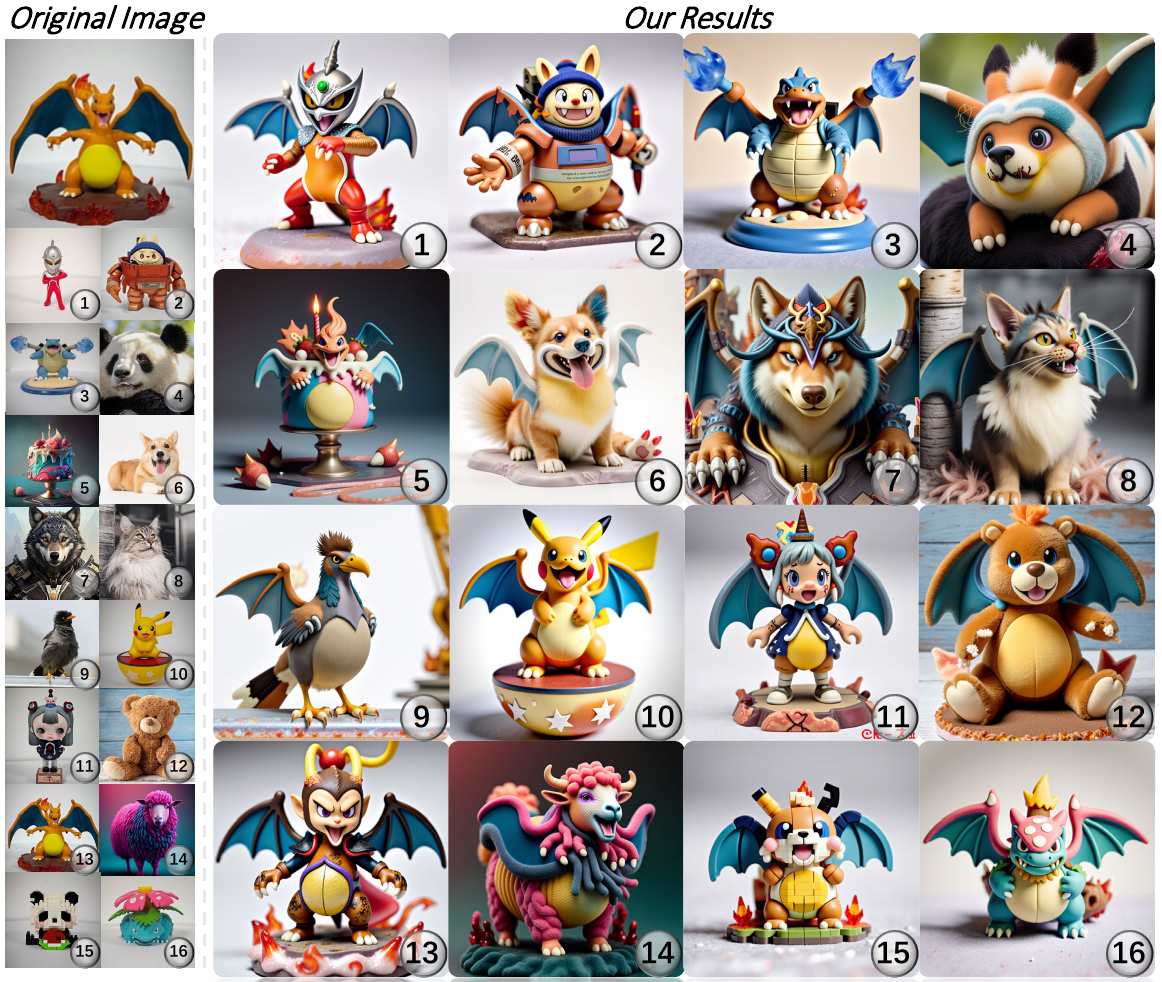}
    \vskip -0.1in
    \caption{\textbf{Additional Qualitative results.} The primary source (\textit{charizard  figurine}, top-left) is fused with secondary inputs (left column), with results shown on the right.}
    \label{fig:More_result_3}
\end{figure}

\begin{figure}[h]
    \centering
    \includegraphics[width=0.87\linewidth]{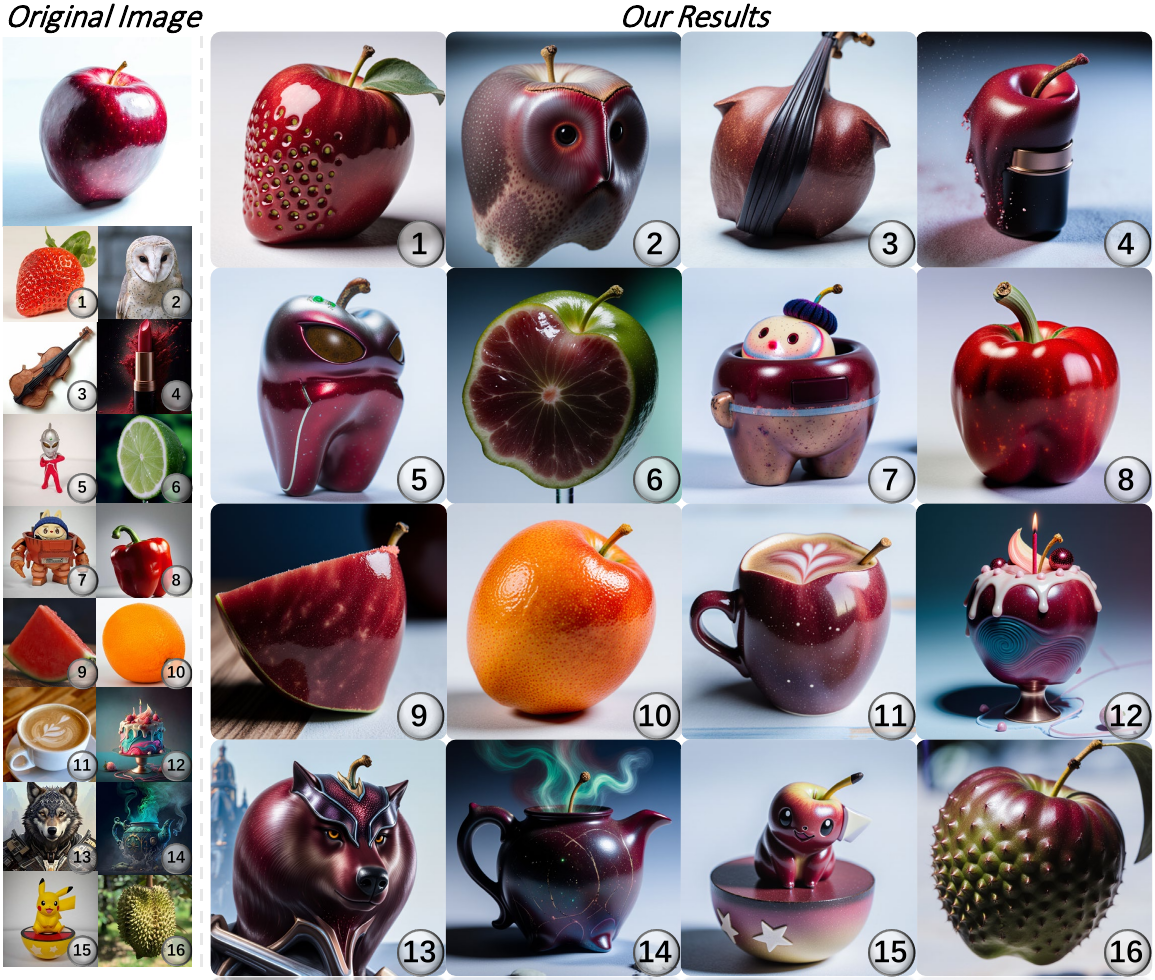}
    \vskip -0.1in
    \caption{\textbf{Additional Qualitative results.} The primary source (\textit{apple}, top-left) is fused with secondary inputs (left column), with results shown on the right.}
    \label{fig:More_result_4}
\end{figure}

\begin{figure}[h]
    \centering
    \includegraphics[width=0.87\linewidth]{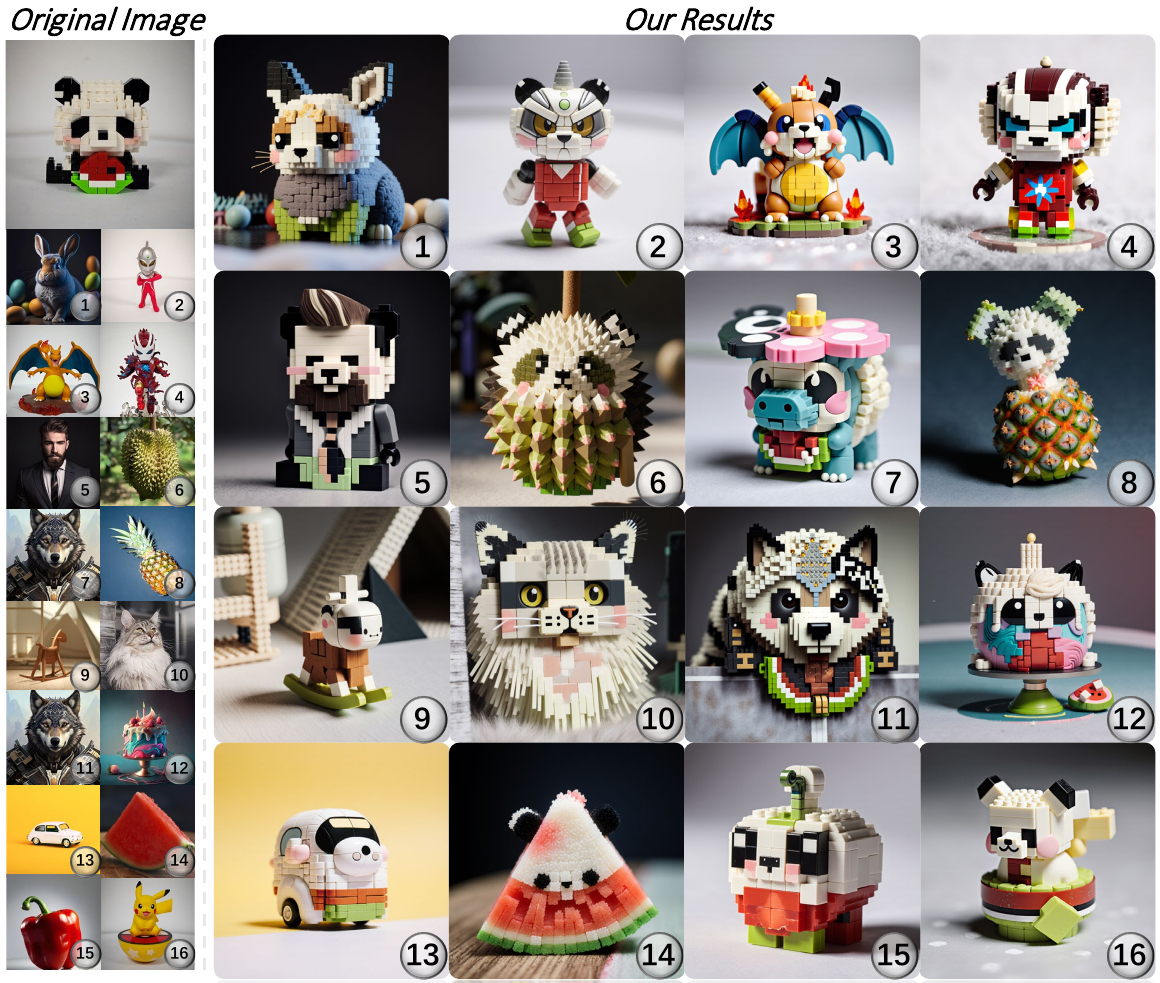}
    \vskip -0.1in
    \caption{\textbf{Additional Qualitative results.} The primary source (\textit{panda figurine}, top-left) is fused with secondary inputs (left column), with results shown on the right.}
    \label{fig:More_result_5}
\end{figure}

\begin{figure}[h]
    \centering
    \includegraphics[width=0.87\linewidth]{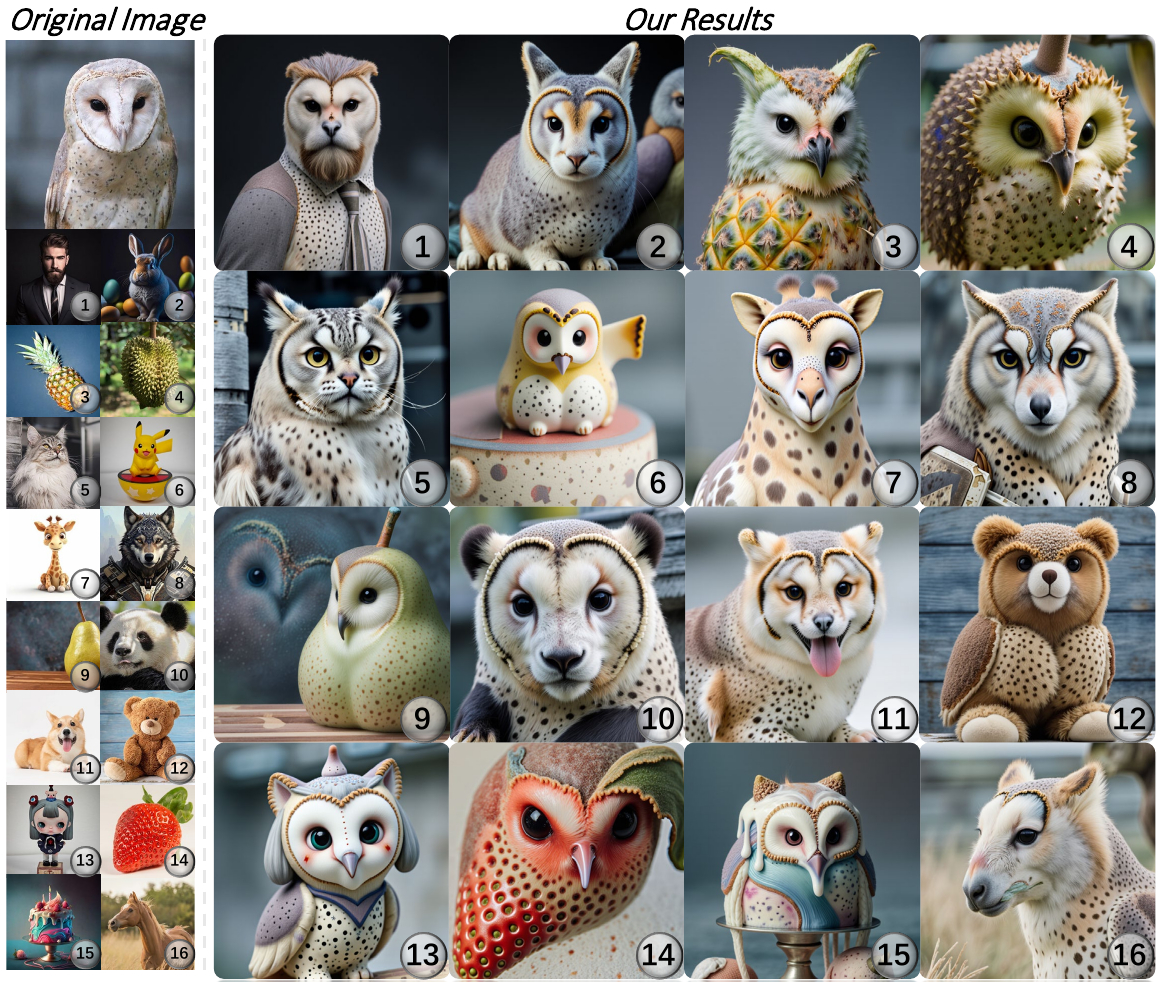}
    \vskip -0.1in
    \caption{\textbf{Additional Qualitative results.} The primary source (\textit{owl}, top-left) is fused with secondary inputs (left column), with results shown on the right.}
    \label{fig:More_result_6}
\end{figure}

\begin{figure}[h]
    \centering
    \includegraphics[width=0.87\linewidth]{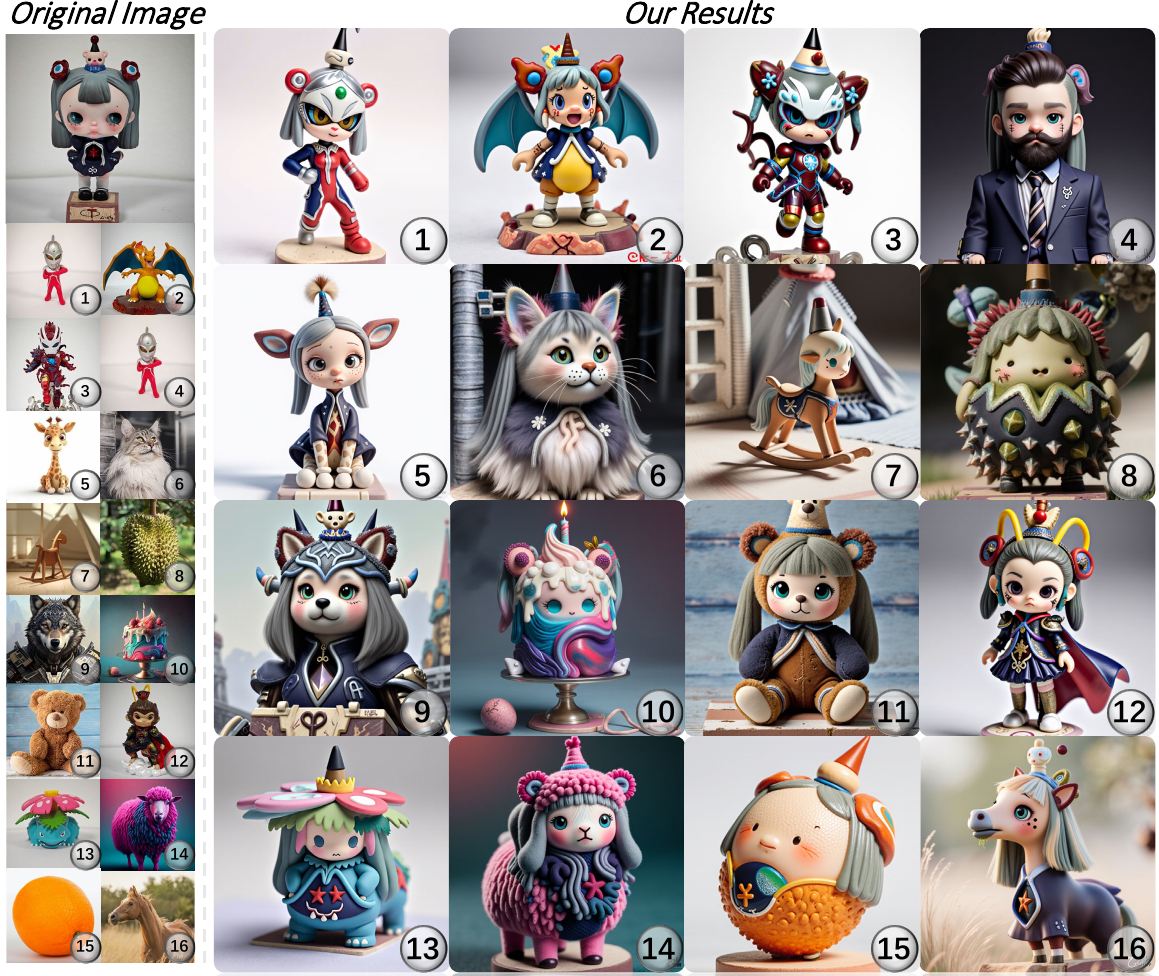}
    \vskip -0.1in
    \caption{\textbf{Additional Qualitative results.} The primary source (\textit{doll figurine}, top-left) is fused with secondary inputs (left column), with results shown on the right.}
    \label{fig:More_result_7}
\end{figure}

\begin{figure}[h]
    \centering
    \includegraphics[width=0.87\linewidth]{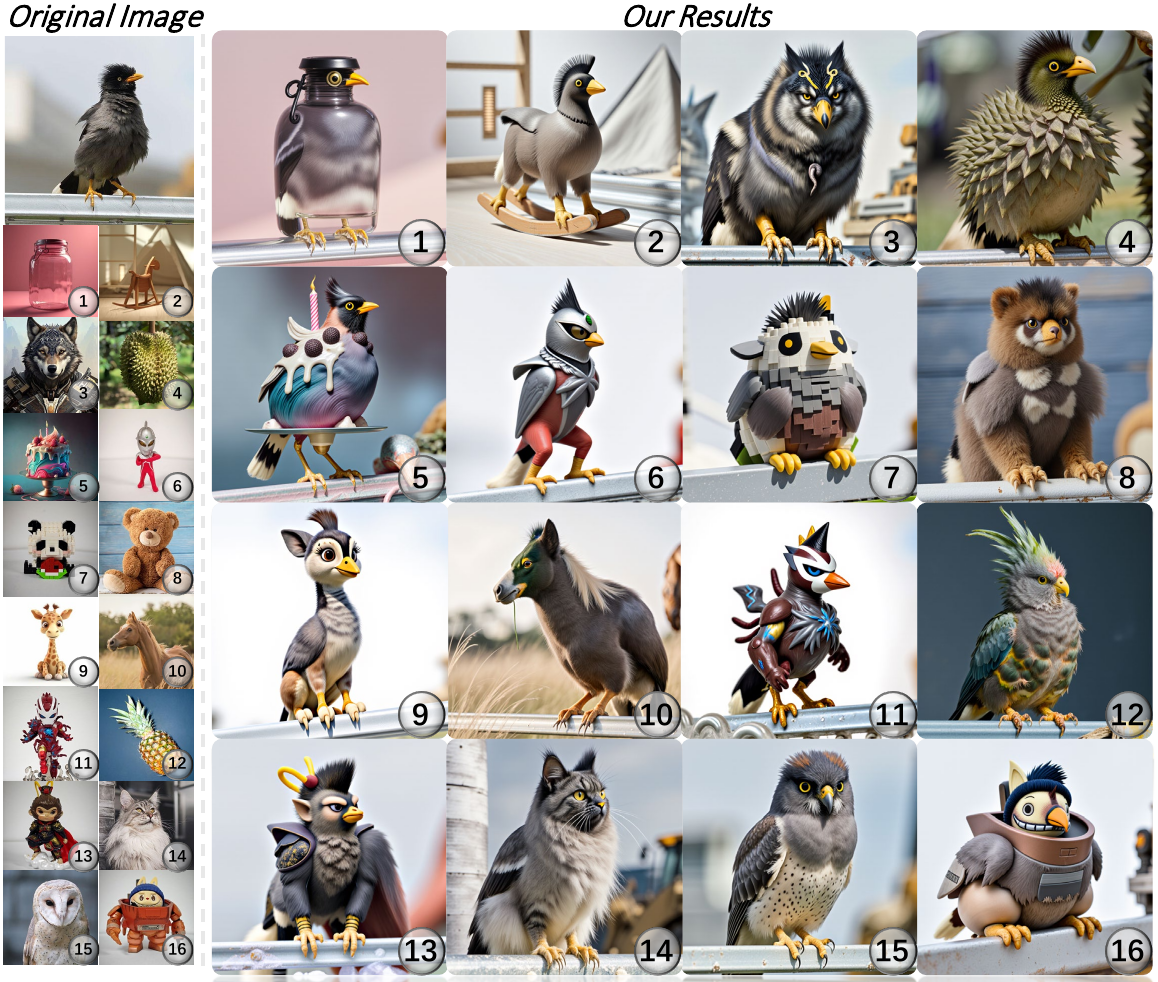}
    \vskip -0.1in
    \caption{\textbf{Additional Qualitative results.} The primary source (\textit{bird}, top-left) is fused with secondary inputs (left column), with results shown on the right.}
    \label{fig:More_result_9}
\end{figure}

\begin{figure}[h]
    \centering
    \includegraphics[width=0.87\linewidth]{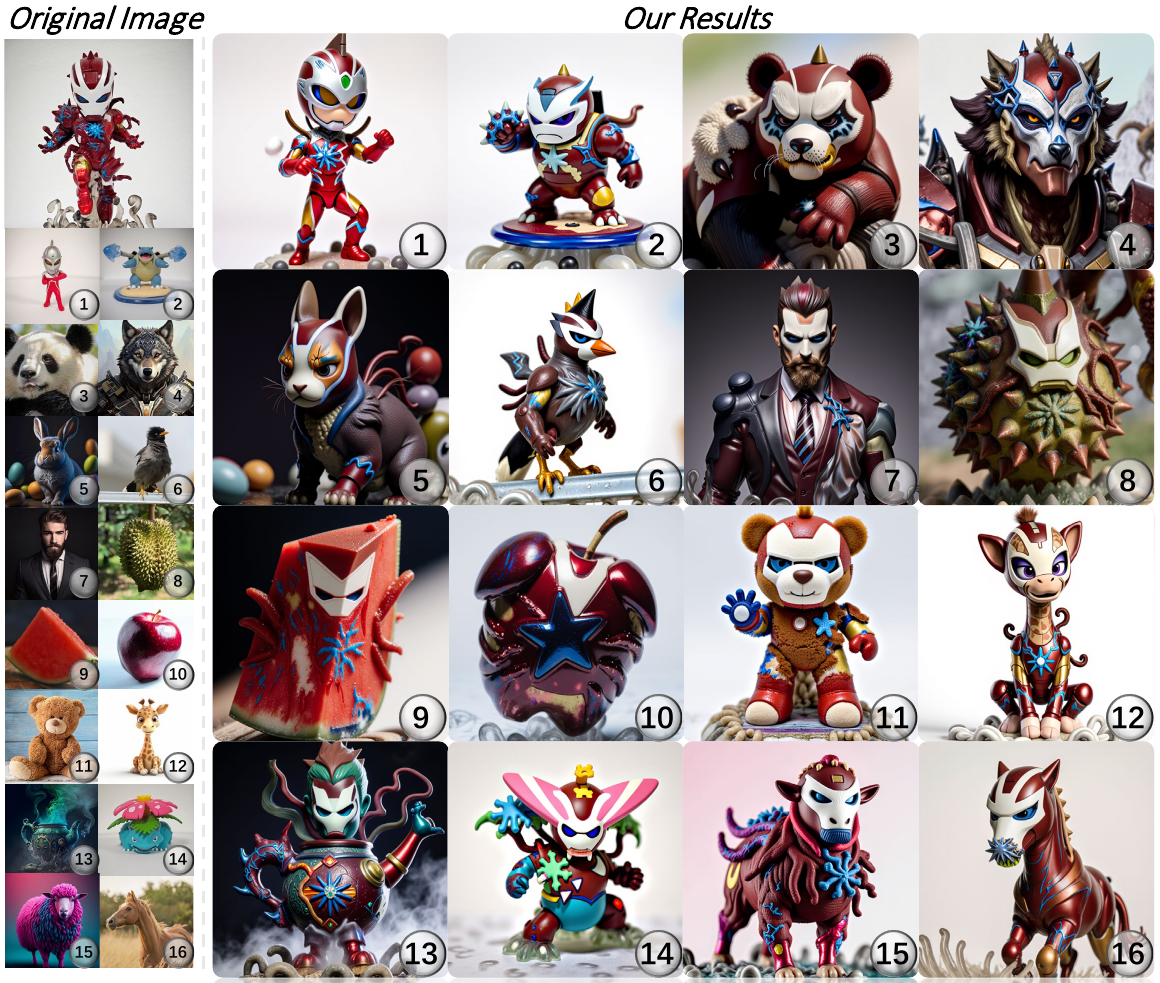}
    \vskip -0.1in
    \caption{\textbf{Additional Qualitative results.} The primary source (\textit{Iron man figurine}, top-left) is fused with secondary inputs (left column), with results shown on the right.}
    \label{fig:More_result_8}
\end{figure}

In this section, we present additional qualitative results with \textbf{resampling disabled}, to evaluate VMDiff under a deterministic setting and further demonstrate its effectiveness and generalization. Fig.~\ref{fig:First_image} shows generations at \(1024\times1024\) resolution. Figs.~\ref{fig:More_result_1}, \ref{fig:More_result_2}, \ref{fig:More_result_3}, \ref{fig:More_result_4}, \ref{fig:More_result_5}, \ref{fig:More_result_6}, \ref{fig:More_result_7}, 
\ref{fig:More_result_9}, and \ref{fig:More_result_8} provide diverse fusion examples spanning animals, fruits, artificial objects, and character figurines. In all figures, the leftmost column displays the source images, and the adjacent columns show the fused outputs.

These examples are generated from our IIOF dataset and cover a wide range of visual appearances and semantic attributes. Across varied fusion types—such as person–fruit, animal–object, and object–object—the results consistently exhibit structural coherence, balanced integration, and high visual fidelity. This indicates that VMDiff can integrate symbolic and structural cues into stylistically consistent hybrids, regardless of whether the source concepts are semantically similar or dissimilar.

Overall, these results substantiate the strong generalization of VMDiff, yielding novel, imaginative, and structurally plausible hybrid objects from diverse real-world inputs, even without resampling or seed variation.

\end{document}